\PassOptionsToPackage{table,dvipsnames}{xcolor} 
\documentclass[11pt]{article}
\usepackage[]{coling}

\usepackage{times}
\usepackage{latexsym}
\usepackage[T1]{fontenc}
\usepackage[utf8]{inputenc}
\usepackage{microtype}
\usepackage{inconsolata}
\usepackage{graphicx}
\usepackage{subcaption}
\usepackage{times}
\usepackage{setspace}
\usepackage{amsmath,amsthm,amsfonts,amssymb,bm}
\usepackage{latexsym}
\usepackage{hyperref}
\usepackage{inconsolata}
\usepackage{url}
\usepackage{adjustbox} 
\usepackage{listings}
\usepackage{multirow}
\usepackage{arydshln} 
\usepackage{enumitem}
\usepackage{caption}
\usepackage{float}
\usepackage{tabularx}
\usepackage{booktabs}
\usepackage{multirow}
\usepackage{tabularx}
\usepackage{array} 
\usepackage{amsmath}
\usepackage{adjustbox}
\usepackage{booktabs} 
\usepackage{bm}

\title{\includegraphics[height=3ex]{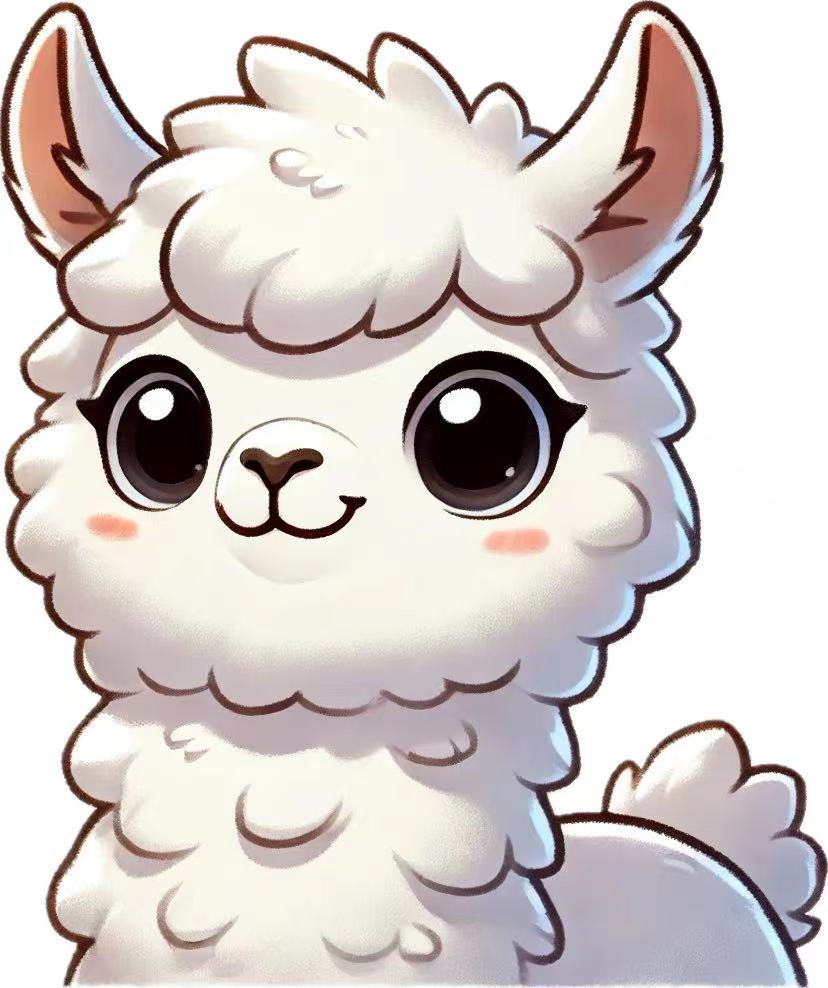} SweetieChat: A Strategy-Enhanced Role-playing Framework for Diverse Scenarios Handling Emotional Support Agent}

\author{
    Jing Ye$^{1,2}$, 
    Lu Xiang$^{1,2}$\Thanks{ Corresponding Author},
    Yaping Zhang$^{1,2}$,
    Chengqing Zong$^{1,2}$\\
    \footnotesize${}^1$State Key Laboratory of Multimodal Artificial Intelligence Systems, Institute of Automation, CAS, Beijing, China\\
    \footnotesize${}^2$School of Artificial Intelligence, University of Chinese Academy of Sciences, Beijing, China\\ 
    \footnotesize{yejing2022@ia.ac.cn}; \footnotesize{\{lu.xiang, yaping.zhang,cqzong\}@nlpr.ia.ac.cn} \\
}

\begin{document}

\maketitle
\begin{abstract}

Large Language Models (LLMs) have demonstrated promising potential in providing empathetic support during interactions. However, their responses often become verbose or overly formulaic, failing to adequately address the diverse emotional support needs of real-world scenarios. To tackle this challenge, we propose an innovative strategy-enhanced role-playing framework, designed to simulate authentic emotional support conversations. Specifically, our approach unfolds in two steps: (1) Strategy-Enhanced Role-Playing Interactions, which involve three pivotal roles---Seeker, Strategy Counselor, and Supporter---engaging in diverse scenarios to emulate real-world interactions and promote a broader range of dialogues; and (2) Emotional Support Agent Training, achieved through fine-tuning LLMs using our specially constructed dataset.
Within this framework, we develop the \textbf{ServeForEmo} dataset, comprising an extensive collection of 3.7K+ multi-turn dialogues and 62.8K+ utterances. We further present \textbf{SweetieChat}, an emotional support agent capable of handling diverse open-domain scenarios. Extensive experiments and human evaluations confirm the framework's effectiveness in enhancing emotional support, highlighting its unique ability to provide more nuanced and tailored assistance.

\end{abstract}

\section{Introduction}

\begin{figure}[t!]
    \centering
    \includegraphics[width=1\linewidth]{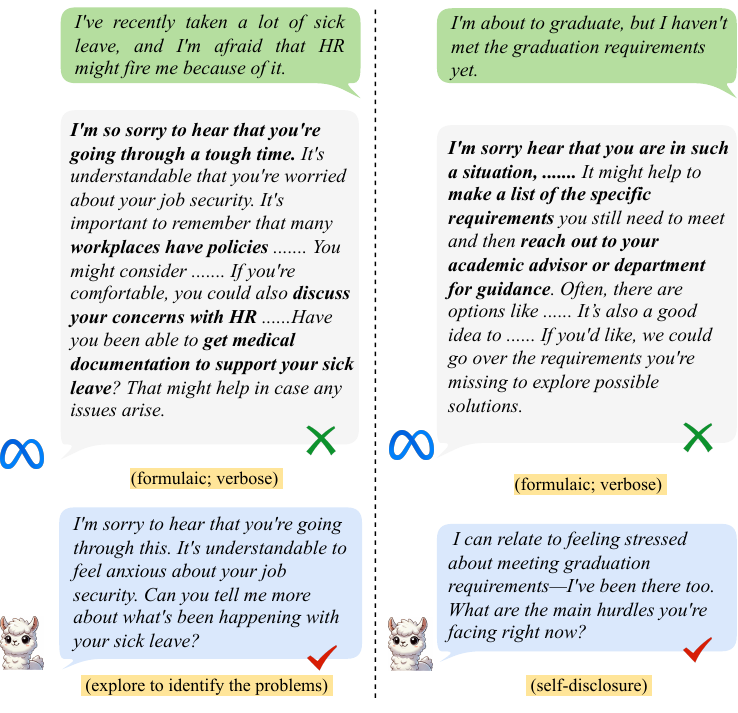}
    \caption{Example responses from the Meta-LLaMA3.1-70B-Instruct and our SweetieChat. LLMs often give verbose and formulaic responses characterized by empathy + suggestions, resulting in a distinct `\textit{AI flavor}'. Conversely, SweetieChat excels in empathy and supportiveness, skillfully addressing and responding to user emotion needs.}
    \label{fig:demo}
\end{figure}
  
Emotional Support Conversation (ESC) systems are designed to alleviate users' emotional distress and help them understand and work through the challenges that they face \cite{ijcai/00080XXSL22, rains2020support}, which play a crucial role in various scenarios, including social interactions, mental health support, and customer service communications. Despite the significant potential of Large Language Models (LLMs) in generating empathetic responses \cite{llama2, qwen2, gpt-4}, current LLMs often struggle to deliver diverse and contextually appropriate support. As depicted in Figure \ref{fig:demo}, the responses of LLMs tend to be formulaic and verbose, lacking the nuanced empathy and tailored suggestions required in real-world applications, which may produce a counterproductive support effect \cite{Muffin, kang-etal-2024-large}.

To enhance the efficacy of ESC systems, the focus has shifted towards crafting high-quality ESC datasets that can fine-tune LLMs to produce more empathetic and context-aware responses \cite{ExTES, ESCoT}. However, manually constructing a high-quality, multi-turn ESC dataset is extremely challenging, which limits the scale and diversity of emotional support scenarios. \citet{ESConv} sourced the ESConv dataset, which only includes 1.3K dialogues spanning 13 topic categories. 

Recent studies leverage the intrinsic generalization capabilities of LLMs to expand and enrich ESC datasets \cite{ExTES, ESCoT}. These initiatives typically employ LLMs to continue, rewrite or imitate existing datasets. Nonetheless, these augmented datasets primarily encounter two significant challenges: 
1) \textit{Insufficient Conversation Diversity}. Although some efforts have notably amplified the dataset volume, the homogeneity among different dialogues remains pronounced \cite{AugESC, ExTES}. This lack of variety can lead to repetitive and predictable interactions, which is hard to meet the unique needs of different seekers. 
2) \textit{Ineffective Strategy Implementation}. In psychology, effective emotional support conversations typically require a strategic approach, involving inquiries to gain deeper insights into the user's situation, thereby enabling more targeted support \cite{strategy, misc}. However, LLMs tend to favor strategies such as Offering Hope, Affirmation and Reassurance, and Providing Suggestions. This tendency results in a deficiency in their empathetic capabilities, as they often overlook the nuanced strategies necessary for truly understanding and assisting users in emotional support interactions.

To address these limitations, we introduce a novel strategy-enhanced role-playing framework for generating emotional support dialogues, as illustrated in Figure \ref{fig:overview}. This framework features three key roles: Seeker, Strategy Counselor, and Supporter. These roles simulate real-world emotional support interactions, enhancing both the diversity of dialogues and the effectiveness of the support provided. Specifically, the Supporter interacts with Seekers from diverse backgrounds, while the Counselor provides guidance by offering well-reasoned and effective strategies to the Supporter. Within this framework, we develop \textbf{ServeForEmo}, an augmented dataset that encompasses a wide range of ES scenarios, featuring 3.7K+ dialogues and 62.8K+ utterances --- approximately 3$\times$ the scale of ESConv. Moreover, we build our emotional support dialogue system, \textbf{SweetieChat}, by fine-tuning LLaMA \cite{llama3} on ServeForEmo. Experimental results confirm that SweetieChat excels in managing diverse out-of-domain scenarios. Our main contributions can be summarized as follows:
\begin{itemize}[itemsep= 0.1pt,topsep = 0.1pt,partopsep=0.1pt]
\item We present a simple yet effective role-playing framework that leverages psychological support strategies to simulate diverse emotional support interactions, enhancing the diversity and rationality of conversations. 
\item We develop ServeForEmo, an efficient ESC dataset that contains a broad spectrum of scenarios, and is supported by valid and effective emotional support strategies.
\item Experimental results and Comprehensive human evaluations demonstrate that our SweetieChat excels in providing emotional support across diverse and open-domain scenarios.
\end{itemize}

\section{Related Work}

\subsection{Emotional Support Conversation}
Emotional support is a crucial capability in human-computer interactions \cite{10.1145/3383123}. 
Although LLMs have revolutionized this field, they often struggle to provide supportive responses. In order to improve the emotional supportive ability of Chatbots, a real-world and comprehensive corpora is of great importance \cite{sharma-etal-2020-computational}. \citet{ESConv} introduce the first manually curated emotional support conversation dataset. However, its size and scope are limited due to the significant expertise and time investment required for its creation. 

Recent efforts have focused on utilizing LLMs to augment existing datasets, with efforts typically falling into two categories: modifying and expanding datasets. \citet{SMILE} approach data augmentation as a rewriting task, creating the \textsc{SMILECHAT} dataset by converting single-turn PsyQA \cite{SunLZLH21} interactions into multi-turn dialogues by using ChatGPT \cite{chatgpt}.  \citet{AugESC} treat data augmentation as a dialogue completion task, where they fine-tune a dialogue LM on ESConv and then prompt it to complete dialogues with posts from the EmpatheticDialogues dataset \cite{RashkinSLB19}. Additionally, \citet{ExTES} expand datasets by leveraging LLMs to imitate and generate new data based on existing datasets. \citet{ESCoT} propose the ESCoT dataset, which improves explainability by adding emotion and strategy reasoning, creating a chain-of-thought in dialogues. While these approaches have augmented a significant amount of data, they often encounter challenges with limited diversity or insufficient strategic guidance.

\subsection{Role-playing}

\begin{figure*}
    \centering
    \includegraphics[width=1\linewidth]{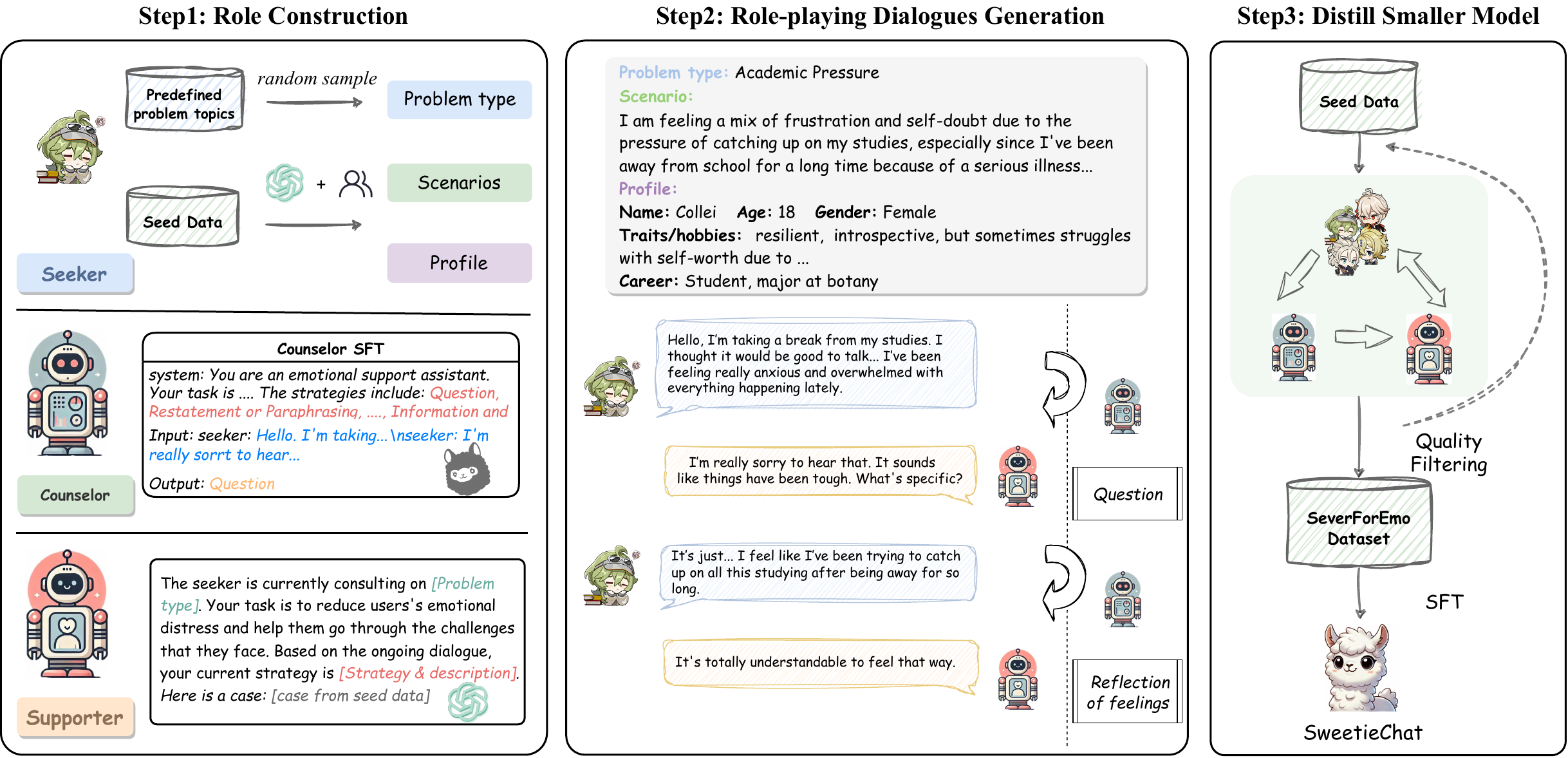}
    \caption{Overview of the proposed strategy-enhanced role-playing framework. Our framework incorporates 3 roles: Seeker, Strategy Counselor, and Supporter. We employ LLMs to simulate these roles and interact with each other like real-world emotional support conversations. Following the dialogue generation process, we develop the ServeForEmo dataset. Building on these foundations, we present SweetieChat, an emotional support agent capable of handling diverse open-domain scenarios.}
    \label{fig:overview}
\end{figure*}

Recent advancements in LLMs have significantly enhanced the development of Role-Playing Agents \cite{CAMEL,RoleLLM, Lu0ZZ24,chen2024personapersonalizationsurveyroleplaying}. These frameworks excel in personalized creation and generation tasks. For example, \citet{SimsChat} propose a role-playing framework to build customizable persona-driven dialogue datasets for casual conversations, while \citet{wu-etal-2024-role} explore role-play in creating interactive drama narratives. Additionally, role-playing has also been used in evaluations. \citet{ESC-Eval} introduce an ESC evaluation framework, which employs role-playing agents to interact with ESC models. Inspired by these developments, we find that diverse roles can enhance the diversity of ESC data.

\section{Method}

Our goal is to build an emotional support agent capable of addressing diverse scenarios with effective psychological support strategies. To achieve this, 
we introduce a strategy-enhanced role-playing framework. 
As shown in Figure \ref{fig:overview}, the proposed framework comprises three main phases: role construction (Section \ref{step 1}), role-playing dialogues generation (Section \ref{step 2}), and the construction of our emotional support agent, SweetieChat (Section \ref{step 3}). 

\subsection{Role Construction}
\label{step 1}
There are three key roles: \textbf{Seeker}, Strategy \textbf{Counselor}, and \textbf{Supporter}. The Seeker plays the user seeking emotional support, while the Supporter offers emotional support. The Strategy Counselor functions as the cognitive component of the Supporter, selecting appropriate strategies to enhance the quality of  responses. Interactions between the Seeker and Supporter contribute to the creation of real-like and comprehensive emotional support dialogues. In the following Sections, we will detail the construction of each role. The prompts for each role are detailed in Appendix \ref{prompts}.

\subsubsection{Seeker Construction}
Diversity in dialogues hinges on the variety of Seekers. 
We enrich the persona of the Seeker from three aspects: seeking problem types, specific scenarios, and the Seeker's profile.

\noindent \textbf{Problem type}
To include a broader range of scenarios, we expand the 13 predefined problem types in ESConv to 45. These problem types are categorized into five groups: Emotional and Mental Health Issues,  Interpersonal Relationships, Personal Development, Life and Work Stress, and Behavioral Issues. Details are provided in Appendix \ref{problem types}. Formally, we denote a specific problem type from the predefined problem pool $\mathcal{P}$ as $p$.

\noindent \textbf{Scenario}
Scenarios reflect the specific issues for which seekers seek assistance. To develop high-quality emotional support dialogues, it's crucial to include diverse, realistic psychological counseling scenarios. Following \cite{ESCoT}, we use GPT-4o to expand the scenario range. Specifically, we establish a seed data pool, $\mathcal{S}=\{\mathcal{S}_i\}_{i=1}^M$, by manually selecting high-quality examples from the ESConv dataset \cite{ESConv} and psychological counseling websites. Each seed data is consist of problem type $p_i$, corresponding scenario $s_i$ and a emotional support dialog $D_i$, accumulating 1,000 distinct $\mathcal{S}_i=(p_i, s_i, \mathcal{D}_i)$ triplets. The scenario descriptions are crafted to be longer than 20 words and detail specific events, avoiding vague references to general issues such as relationship breakdowns or mood swings. For a new problem type $p'$, we randomly select a corresponding scenario from $\mathcal{S}$ and use it to generate a new scenario with GPT-4o. 
This process can be formalized as follows:
\begin{align}
    s' & \gets \text{GenerateScenario}(p', \mathcal{S}_i) 
\end{align}

\noindent \textbf{Seeker Profile}
Intuitively, individuals with different backgrounds and personalities often exhibit different reactions to identical scenarios. To enhance the diversity of our dialogues, we craft a tailored seeker profile for each dialogue based on the problem type and specific scenario. Initially, we establish a seed profile pool $\mathcal{C}=\{\mathcal{C}_i\}_{i=1}^{N}$ comprising 100 handcrafted seeker profile. Each profile $\mathcal{C}_i$ comprises a problem type $p_i$, scenario $s_i$, and character description $c_i$. Building on prior work in personality-driven dialogues \cite{zhang-etal-2018-personalizing, abs-1901-09672}, $c_i$ features attributes such as name, gender, address, occupation, personality traits, and hobbies. We leverage the context learning capability of GPT-4o to dynamically generate a unique character $c'$ for each new problem type $p'$ and scenario $s'$. The new profile is then added back to enrich the seed profile pool $\mathcal{C}$: 
\begin{align}
   c' & \gets \text{GenerateProfile}(p', s', \mathcal{C}_i) \\
    \mathcal{C} & \gets \mathcal{C} \cup \{(p', s', c')\}
\end{align}

\subsubsection{Counselor Construction} 
\label{Counselor Construction} 
While LLMs excel at generating empathetic responses, they still struggle provide emotional support using various support skills. There is a significant divergence in strategy preferences between LLMs and human counselors. A typical emotional support conversation consists of three stages: exploration, comforting, and action \cite{Hill}. In contrast, LLMs tend to favor strategies such as offering hope, affirmation and reassurance, as well as providing suggestions \cite{kang-etal-2024-large}. More evidences are in Appendix \ref{counselor training}. This tendency to rush solutions and provide empathetic responses results in a lack of deeper interaction and understanding with the seeker, markedly differing from the more nuanced approach typical of human interactions.

To mitigate these challenges, we utilize the strategy-enhanced LLaMA as our support strategy counselor. We design a strategy-selection task, and then fine-tune LLaMA using the manually annotated ESConv dataset. The Counselor is instructed to select the appropriate emotional support strategy based on the the ongoing dialogue history, as formalized by the equation: 
\begin{equation} 
o_i = \underset{o \in \mathcal{O}}{\operatorname{arg\,max}} \mathcal{P}(o |\mathcal{D}_{i}^j) 
\end{equation}
Here, $\mathcal{O}$ represents the set of strategies, and $\mathcal{D}_i^j$ denotes the dialogue history up to the $j$-th round of dialogue $\mathcal{D}_i$, consisting of the sequence of previous utterances \{$u_i^1, r_i^1, \dots, r_i^{j-1}, u_i^j$\}. Here, $u_i^j$ is the $j$-th statement from the Seeker, and $r_i^{j-1}$ is the most recent response from the Supporter.

\subsubsection{Supporter Construction}

We employ GPT-4o to act as the supporter, known for its proficiency in generating empathetic responses and in-context learning. The supporter's actions are guided by a support strategy selected by the counselor. Given the chosen strategy, along with the dialogue history and specific case dialogues, GPT-4o is tasked with producing emotionally supportive replies. 

\subsection{Role-playing Dialogue Generation}
\label{step 2}
The interactions between seekers from diverse backgrounds and the supporter form the core of our emotional support dialogue. The counselor serves as an assistant, providing well-reasoned support strategies to guide the supporter in delivering clear and interpretable emotional support. \textit{In this scene, our framework not only accommodates diverse scenarios but also ensures reliable strategy guidance.} To enhance the diversity of our data, we implement a self-iterative mechanism to extend the seed data pool\cite{ExTES}. 
\begin{align}
    \mathcal{S} & \gets \mathcal{S} \cup \{(p',s',\mathcal{D'})\}
\end{align}
where $p'$ is the randomly selected problem type, $s'$ represents the newly generated scenarios, and $\mathcal{D'}$ is the simulated emotional support conversation.

\subsection{SweetieChat Agent}
\label{step 3}
Our strategy-enhanced role-playing framework facilitates the development of the ServeForEmo dataset, which captures a wide range of seekers and scenarios. This dataset is designed to narrow the empathy response gap between LLMs and humans. Building upon this foundation, we develop SweetieChat, a specialized agent tailored for providing emotional support. To optimize its performance, we fine-tune the LLaMA model using the ServeForEmo dataset.

\section{Data Analysis}
This section explores key concerns about the ServeForEmo dataset:
\noindent Q1: What is the overall quality of the ServeForEmo dataset? 
\noindent Q2: Does the dataset exhibit sufficient diversity?
\noindent Q3: How are the strategies implemented in the dialogues?

\subsection{Statistics (Q1)}
\begin{table}[htp]
\centering
\small
\setstretch{1.25}
\begin{adjustbox}{width=\columnwidth}
\begin{tabular}{clcc}
\toprule[1.2pt]
\multicolumn{2}{c}{\textbf{Category}}                & \textbf{ESConv} & \textbf{SeverForEmo} \\ \hline
\multirow{4}{*}{Total}     & \# Dialogues            & 1,300            & 3,757                 \\
                           & \# Utterances           & 29,278           & 62,863                \\
                           & Avg. Dialogue Length    & 22.54           & 16.73                \\
                           & Avg. Utterance Length    & 21.17           & 17.97                \\ \hline
\multirow{3}{*}{Seeker}    & \# Utterances           & 14,639           & 30,722                \\
                           & Avg. \# Utter. per Dialog & 11.27           & 8.18                 \\
                           & Avg. Utterance Length    & 19.9            & 15.25                \\ \hline
\multirow{3}{*}{Supporter} & \# Utterances           & 14,639           & 32,141                \\
                           & Avg. \# Utter. per Dialog & 11.27           & 8.55                 \\
                           & Avg. Utterance Length   & 22.45           & 20.56                \\ \bottomrule[1.2pt]
\end{tabular}
\end{adjustbox}
\caption{Comparison of ServeForEmo and ESConv statistics. For ESConv, we exclude initial greetings and merge consecutive utterances from the same speaker to simplify the dialogue.}
\label{tb:statistics}
\end{table}
To guarantee high data quality, we perform thorough post-processing that includes filtering out low-quality dialogues, eliminating redundant greetings, and managing dialogue length. After human review and refinement, we develop the ServeForEmo dataset, which comprises 3.7K+ dialogues and 62.8K+ utterances, approximately 3$\times$ the scale of ESConv. The detailed statistics are presented in Table \ref{tb:statistics}. Since we limit utterances to a maximum of three sentences during role interactions, the dialogues and utterances in ServeForEmo are generally shorter than those in ESConv. For additional details on our dataset post-processing and quality evaluation methods, please refer to Appendix \ref{data quality evaluation}.

\subsection{Diversity Analysis (Q2)}

To illustrate the broad scope of the ServeForEmo dataset, we focus on two key aspects of diversity: semantic and lexical features. For brevity, we only present semantic diversity in the main body of our paper. Details on the other diversity aspect are in Appendix \ref{diversity analysis 2}.

\begin{figure}
    \centering
    \begin{subfigure}[b]{0.48\linewidth}
        \centering
        \includegraphics[width=\linewidth]{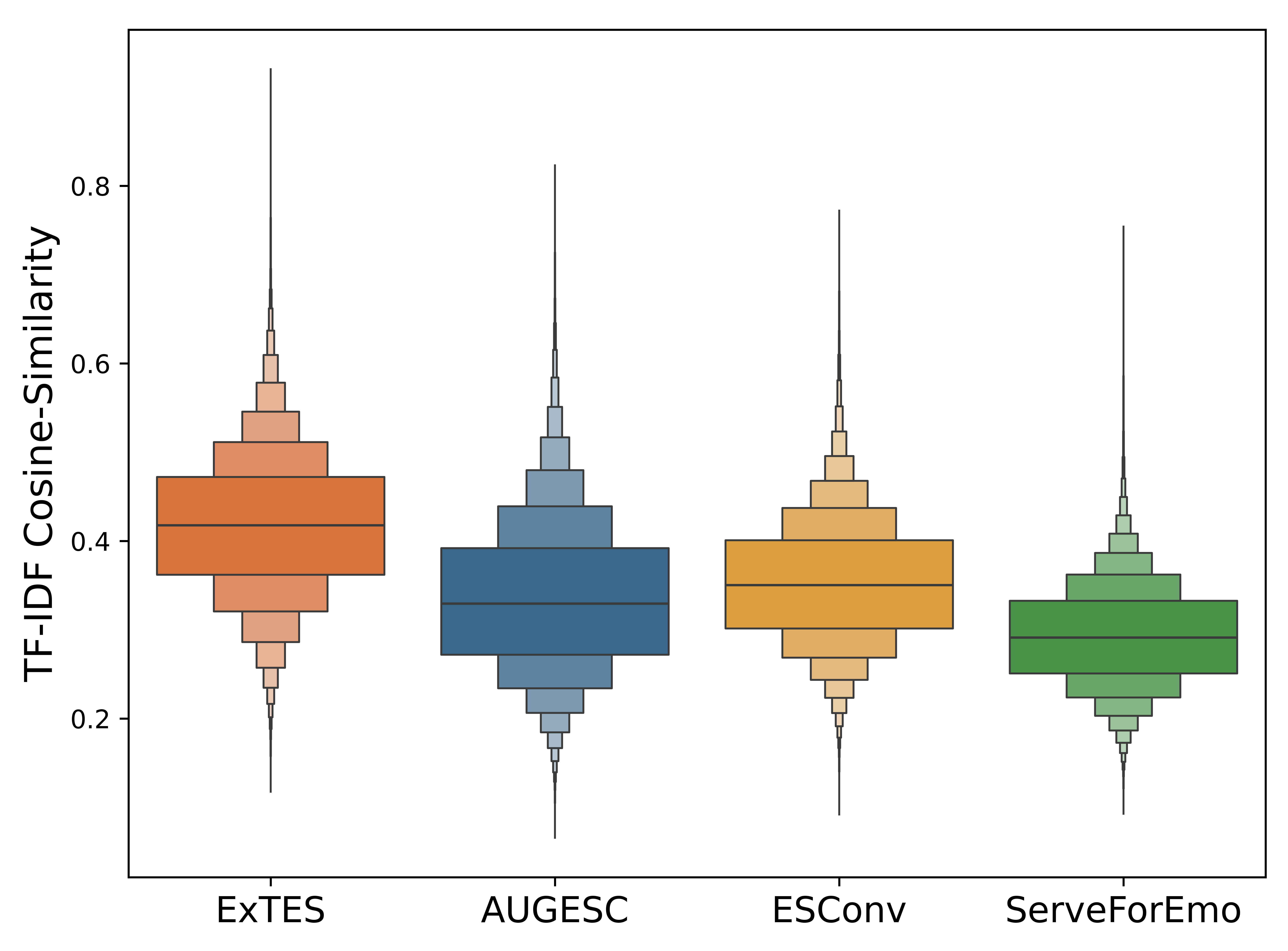} 
        \caption{}
        \label{fig:subfig1}
    \end{subfigure}
    \hfill
    \begin{subfigure}[b]{0.48\linewidth}
        \centering
        \includegraphics[width=\linewidth]{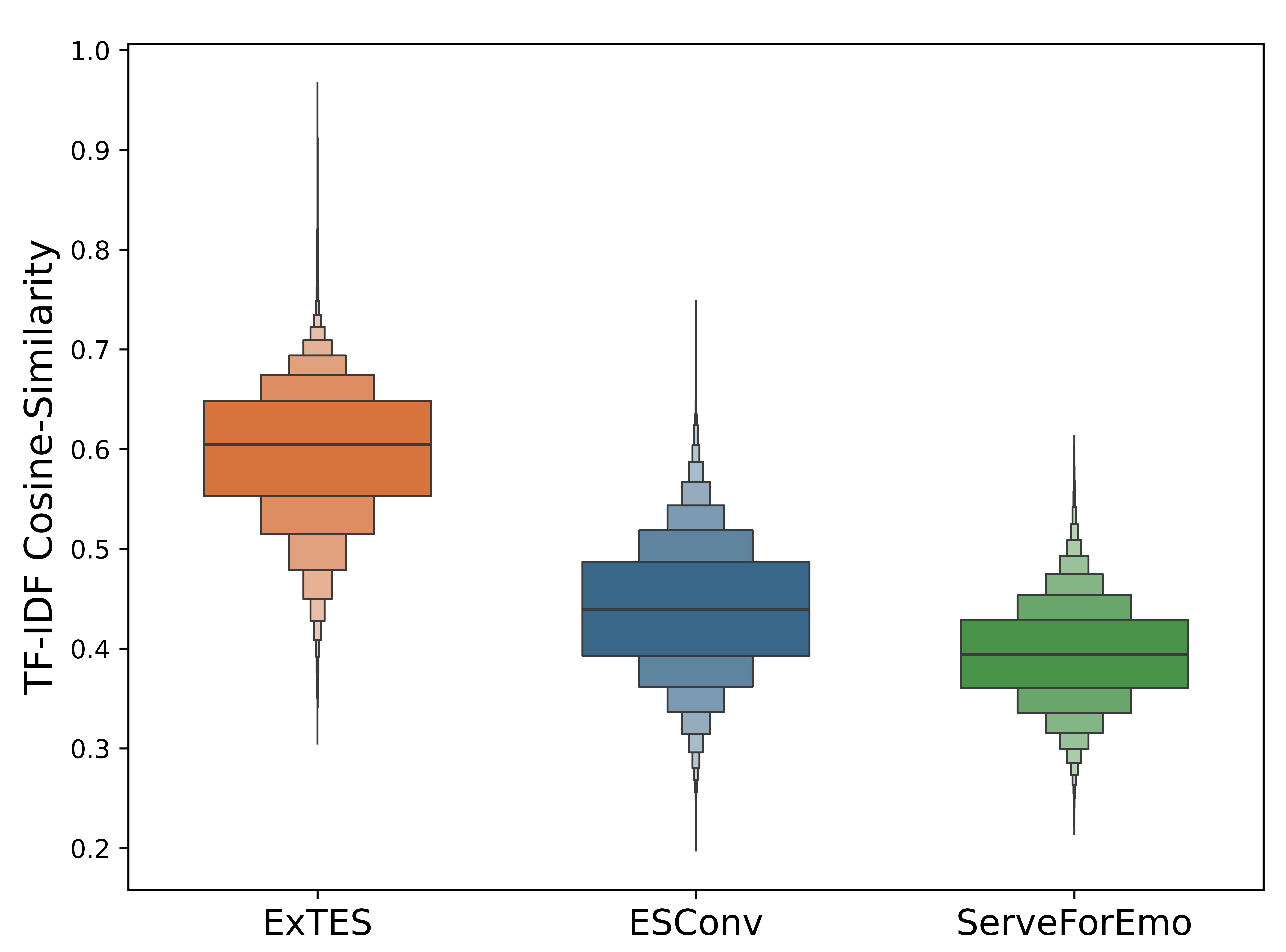}  
        \caption{}
        \label{fig:subfig2}
    \end{subfigure}
    \hfill
    \caption{(a) Global inter-dialogue similarity statistics computed using TF-IDF vectors. 
    (b) Inter-dialogue similarity statistics within the academic field. Lower similarity values indicate higher diversity.
    }
    \label{fig:tf-idf}
\end{figure}

\begin{figure}
    \centering
    \begin{subfigure}[b]{\linewidth}
        \centering
        \includegraphics[width=\textwidth]{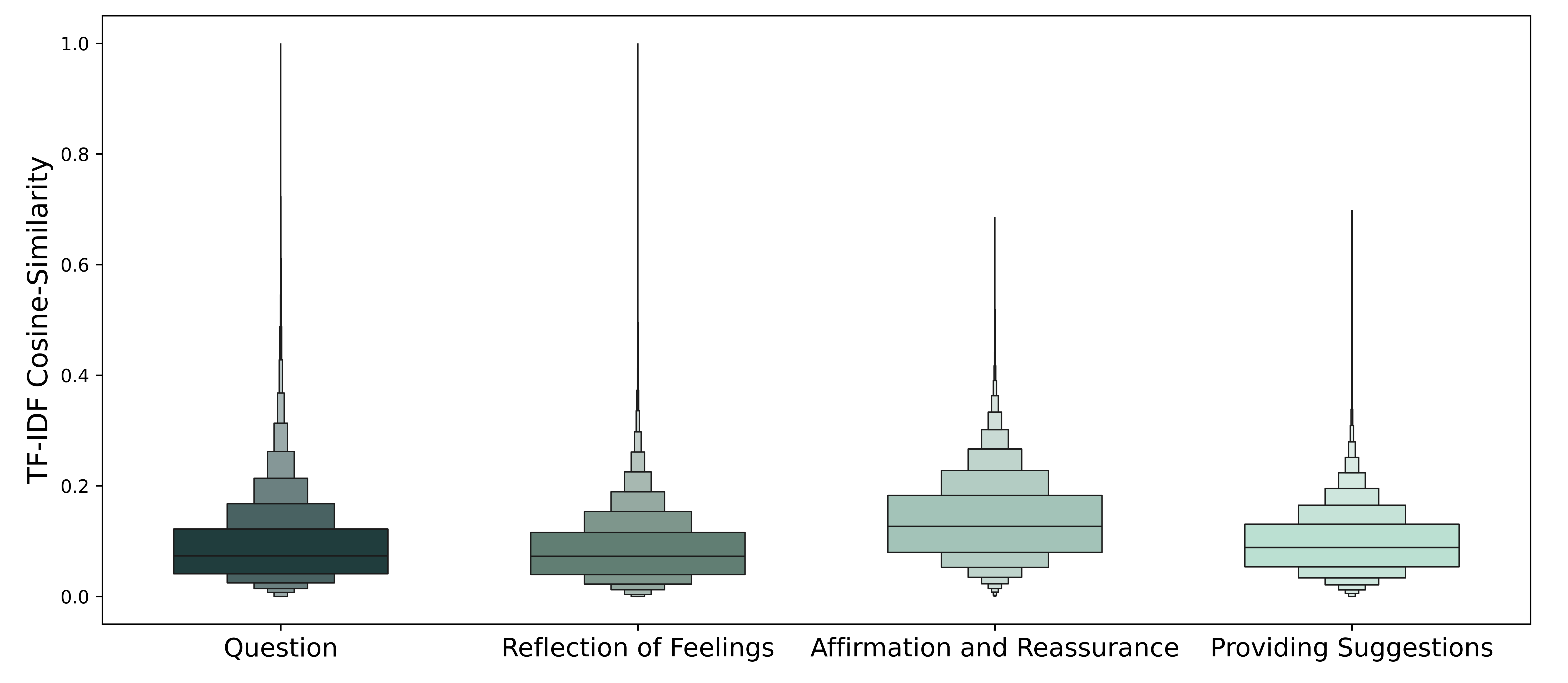} 
        \caption{}
        \label{fig:similarity_full_strategy}
    \end{subfigure}
    \hfill
    \begin{subfigure}[b]{\linewidth}
        \centering
        \includegraphics[width=\textwidth]{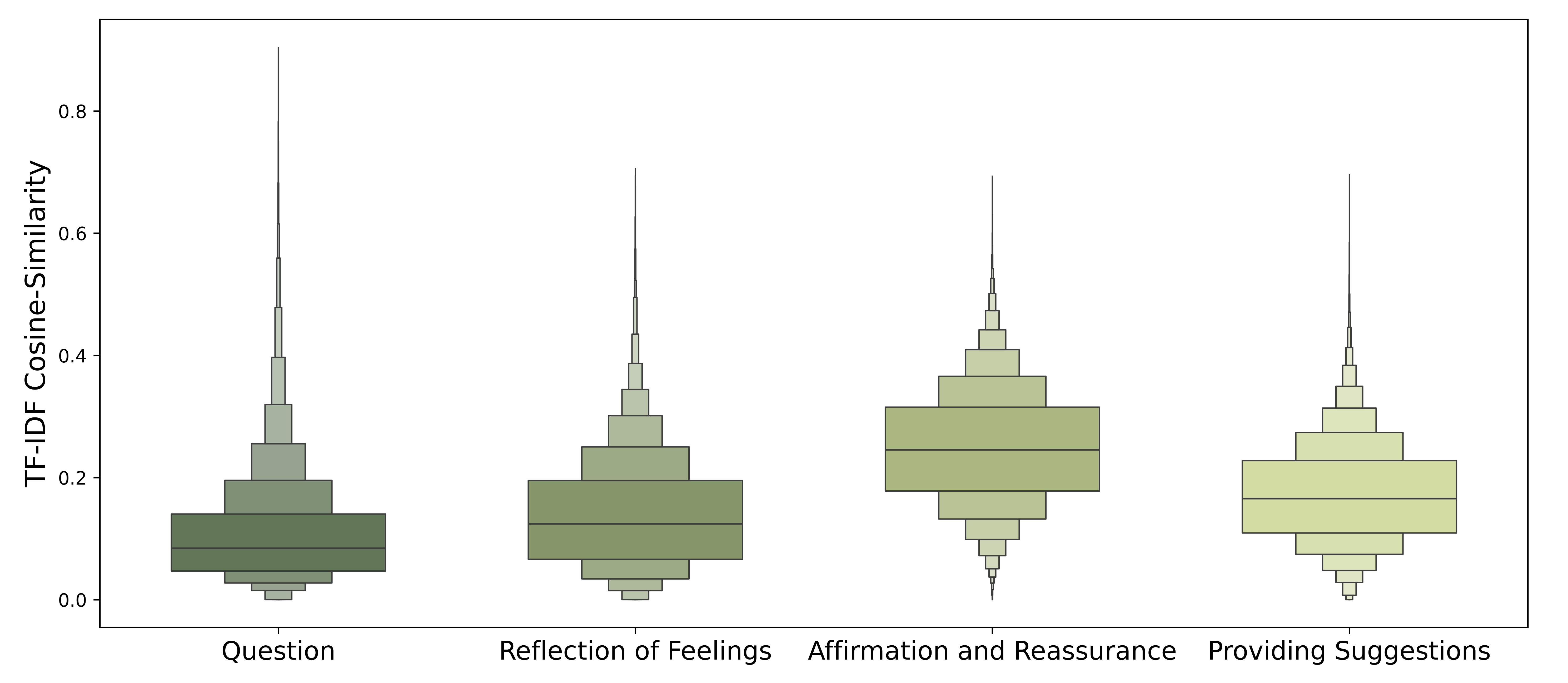}  
        \caption{}
        \label{fig:similarity_academic_strategy}
    \end{subfigure}
    \caption{(a) Global similarity statistics of responses guided by the same support strategy, computed using TF-IDF features. (b) Similarity statistics of responses within the academic field under the same support strategy.}
    \label{fig:strategy_diversity}
\end{figure}

To assess the semantic diversity of ESC datasets, we calculate the cosine similarity between pairs of distinct dialogues using TF-IDF features \cite{tfidf}. Figure \ref{fig:subfig1} and \ref{fig:subfig2} demonstrate that the ServeForEmo dataset exhibits the best inter-dialogue diversity. The low similarity across both global and specific problem-type fields indicates that ServeForEmo effectively captures a wide range of emotional support scenarios. Dialogues collected through our role-playing framework even have higher inter-dialogue diversity compared to the crowdsourced ESConv dataset. Conversely, datasets generated through direct prompting of LLMs, such as ExTES, display minimal diversity. Although these efforts have notably increased the dataset volume, the homogeneity among different dialogues remains pronounced. Since the diversity of the generated data relies heavily on the prompts, it is challenging to produce varied and realistic conversations through prompting.

Despite the diversity of dialogues, we are curious about whether supporters' responses, guided by the same strategies, demonstrate variation across different scenarios. Thus, we analyze the responses associated with the four most frequently applied strategies, and present the similarity statistics of responses under the same strategic guidance in Figure \ref{fig:strategy_diversity}. The results reveal that supporters' responses, even under the same strategy guidance, exhibit low cosine similarity scores. This indirectly confirms that the responses in ServeForEmo are tailored to specific scenarios.

\subsection{Strategy Analysis (Q3)}
\begin{table}[htp]
\centering
\setstretch{1.25}
\begin{adjustbox}{width=\columnwidth}
\begin{tabular}{lcc}
\toprule[1.2pt]
{\textbf{Strategy}} & {\textbf{ESConv}} & {\textbf{ServeForEmo}}  \\ \hline
Question                                               & 20.68\%                          & 19.04\%                               \\
Others                                                 & 18.18\%                          & 6.06\%                                \\
Providing suggestions                                  & 16.08\%                          & 20.00\%                               \\
Affirmation and reassurance                            & 15.38\%                          & 25.26\%                               \\
Self-disclosure                                        & 9.32\%                           & 4.93\%                                \\
Reflection of feelings                                 & 7.81\%                           & 12.03\%                               \\
Information                                            & 6.61\%                           & 8.79\%                                \\
Restatement or paraphrasing                            & 5.93\%                           & 3.90\%                                \\ \bottomrule[1.2pt]
\end{tabular}
\end{adjustbox}
\caption{Strategy distribution of ESConv and ServeForEmo.}
\label{tab:strategy}
\end{table}

\begin{figure}
    \centering
    \begin{subfigure}[b]{\linewidth}
        \centering
        \includegraphics[width=\textwidth]{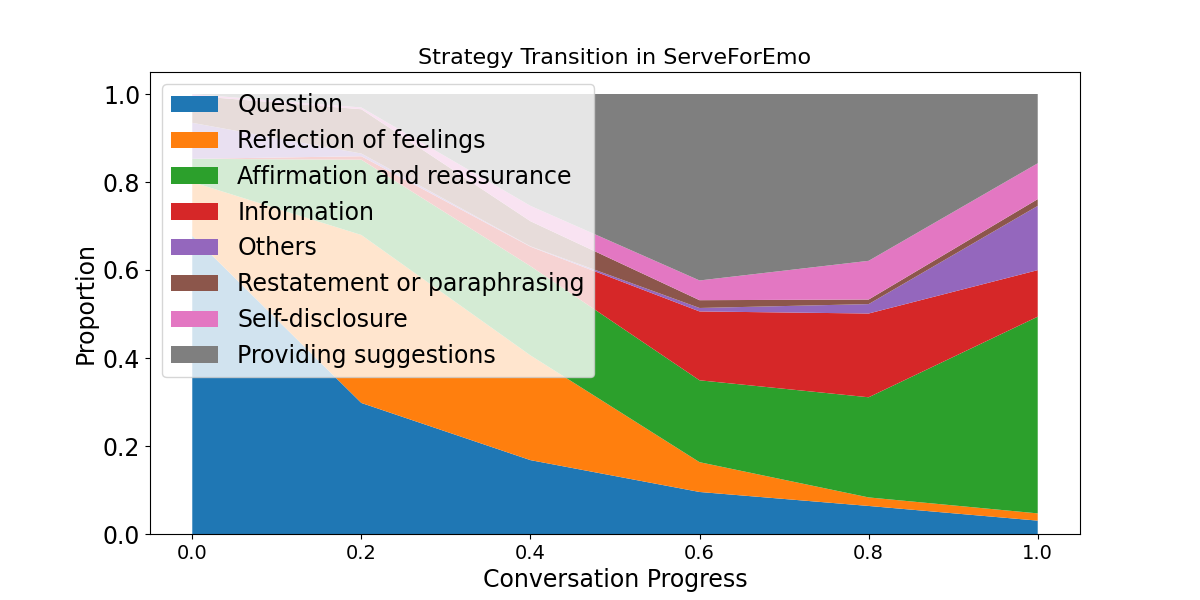} 
        \caption{}
        \label{fig:strategy_transition}
    \end{subfigure}
    \hfill
    \begin{subfigure}[b]{\linewidth}
        \centering
        \includegraphics[width=\textwidth]{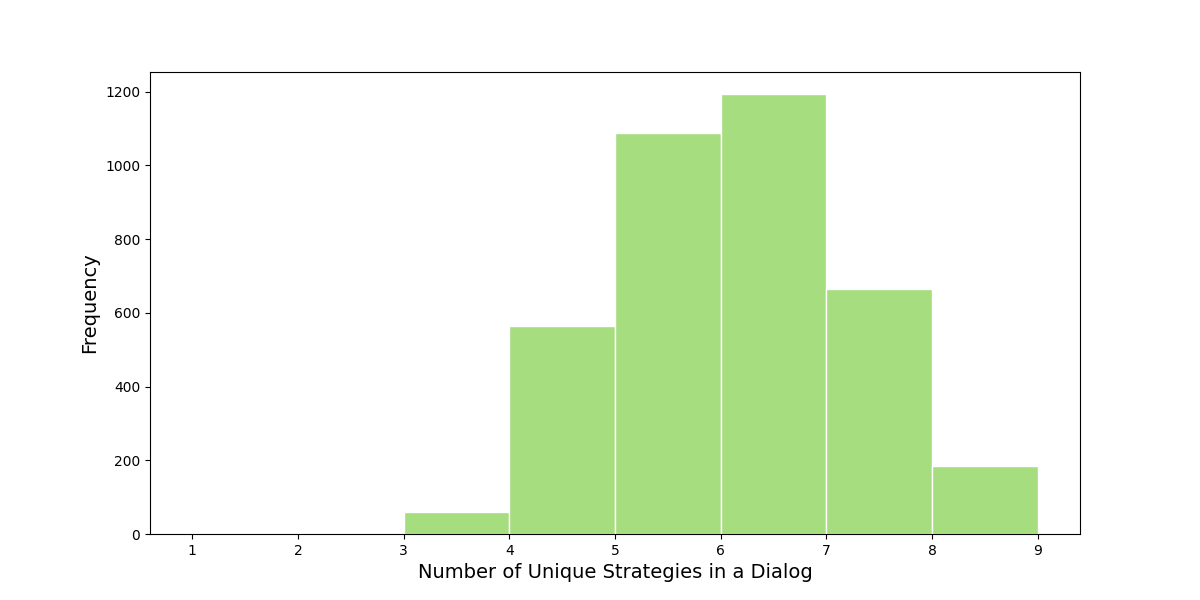}  
        \caption{}
        \label{fig:strategy_counts}
    \end{subfigure}
    \caption{(a) Distribution of strategies at different conversation progress on ServeForEmo dataset. 
    (b) Distribution of unique strategies across different dialogues on ServeForEmo.}
    \label{fig:strategy}
\end{figure}

\noindent \textbf{Strategy Distribution} As described in Section \ref{Counselor Construction}, we perform instruction fine-tuning on LLaMA using the ESConv dataset to be the Strategy Counselor. Table \ref{tab:strategy} presents a comparative analysis of strategy distributions between the ServeForEmo and ESConv datasets. The results reveal a strong similarity between the two datasets, with the main strategies concentrated on Affirmation and Reassurance, Providing Suggestions, and Questioning. This demonstrates the Counselor's effectiveness in selecting and assessing strategies.

\noindent \textbf{Strategy Transition} 
To show whether the strategies annotated by our counselor follow a reasonable procedure, we analyze the transition of strategies across different phases of the conversation. We segment the conversation progress into six intervals and count the proportions of different strategies within each interval for all the conversations in ServeForEmo. Figure \ref{fig:strategy_transition} illustrates the strategy distribution across different conversation progress points. The result shows that in the early stages of the conversation, our counselor primarily employs Questioning to explore and help seekers identify their problems. As the conversation progresses into the middle and later stages, the Providing Suggestions strategy becomes more prominent, assisting seekers in finding solutions. Throughout the conversation, strategies such as Affirmation and Reassurance and Reflection of Feelings are consistently applied, providing comfort by conveying empathy and understanding. This strategic arrangement reflects the three-stage helping skill model proposed by \citet{Hill}, indicating that the counselor is capable of effectively organizing emotional support strategies, making it a competent assistant.

\noindent \textbf{Distribution of Unique Strategies in Dialogues} Here, we aim to show the diversity of strategies employed throughout our conversations. Figure \ref{fig:strategy_counts} presents the distribution of unique strategies within individual dialogues in ServeForEmo. Most dialogues feature more than five distinct strategies, underscoring the wide range of emotional support techniques employed in the dataset. This diversity aligns with our goal of providing comprehensive emotional support through diverse counseling strategies.

\section{Experiments}
\subsection{Baseline Models}

To validate the effectiveness of the ServeForEmo Dataset, we conduct a comprehensive comparison, detailed below:

\noindent \textbf{LLaMA} \cite{llama3}: A foundational open-source LLM. For our experiments, we employ the \texttt{LLaMA3-8B-Instruct} version as the base model, incorporating task-specific settings and applicable strategies within the system prompts.

\noindent \textbf{AUGESC }\cite{AugESC}: Fine-tuned from LLaMA using the AUGESC dataset, which is crafted by leveraging a fine-tuned dialogue LM to do dialogue completion task.

\noindent \textbf{ExTES }\cite{ExTES}: Fine-tuned from LLaMA using the ExTES dataset, which is generated by prompting LLMs to imitate seed conversations.

\noindent \textbf{ESConv }\cite{ESConv}: Fine-tuned from LLaMA using the ESConv dataset. The meticulous human curation of ESConv ensures that the model is trained on dialogues rich in real-life emotional content and effective emotional support responses.

\noindent \textbf{ServeForEmo (ours)}: Also known as SweetieChat, this model is specifically fine-tuned from LLaMA using our ServeForEmo dataset, aiming to enhance its performance in diverse ES scenarios.

\subsection{Implementation Details}
\label{training setting}
We apply LoRA adaptation \cite{LoRA} to the $W_q$, $W_v$, $W_k$, and $W_o$ parameters, with a rank ($r$) set to 8. For optimization, we use the AdamW optimizer \cite{adamw}, setting a learning rate of $3 \times 10^{-5}$ and employing a linear warm-up over the first 1\% of total training steps. The per device batch size is set to 2, and the training is conducted over 5 epochs. To prevent overfitting, we implement early stopping with a patience of 7 evaluation steps. The model achieving the best performance on the validation set is selected for testing. All experiments are conducted on eight Tesla V100 GPUs, and the final results are reported as the average of three experimental runs.

\subsection{Evaluation Settings}
\label{evaluation settings}
To ensure equitable comparison across different models, we employ the manually annotated test set of ESConv as our benchmark. This allows us to assess the performance of models under real-world human annotation conditions. Our evaluation metrics are categorized into two types: automatic and human.

\noindent \textbf{Automatic Evaluation Metrics}
We employ 4 established automatic evaluation metrics: BLEU-n \cite{bleu}, ROUGE-L \cite{rouge}, Distinct-n \cite{diversity}, and BERT-Score \cite{BERTScore}. The responses are tokenized using the LLaMA3 \cite{llama3} Tokenizer. Each metric quantitatively assesses distinct aspects of textual quality, providing a comprehensive overview of model performance.

\noindent \textbf{Human Evaluation Metrics}
For human evaluation, we focus on evaluating the generated responses. Specifically, we focus on the following dimensions: Empathy, Informativeness, Coherence, Suggestion, Understanding, Helpfulness, and Overall Quality. A detailed description of these evaluation metrics is provided in Appendix \ref{human metrices}.

\subsection{Main Results}

\subsubsection{Automatic Evaluation Results}
\begin{table*}[t]
\centering
\footnotesize
\setstretch{1.25}
\begin{adjustbox}{width=\textwidth}
\begin{tabular}{ccccccccc}
\toprule[1.2pt]
\textbf{Model} & \textbf{BLEU-2}      & \textbf{BLEU-4}      & \textbf{R-2}         & \textbf{R-L}          & \textbf{Distinct2}    & \textbf{Distinct3}         & \textbf{BERT-Score} \\ \hline
\multicolumn{8}{c}{\textit{\textbf{Interactive   evaluation with generated context}}}                                                                                                                                 \\ \hline
ESConv         & 6.15$_{\pm 0.05}$         & 2.95$_{\pm 0.06}$          & 3.07$_{\pm 0.10}$          & 13.87$_{\pm 0.15}$          & 98.15$_{\pm 0.89}$          & 98.40$_{\pm 0.82}$       & 85.49$_{\pm 0.10}$           \\ \cdashline{1-8}
LLaMA          & 3.62$_{\pm 0.09}$          & 1.28$_{\pm 0.04}$          & 2.33$_{\pm 0.11}$          & 10.41$_{\pm 0.18}$         & 94.75$_{\pm 0.09}$          & 98.71$_{\pm 0.01}$       & 83.80$_{\pm 0.06}$            \\
ExTES          & 5.92$_{\pm 0.06}$          & 2.44$_{\pm 0.06}$          & \textbf{2.74$_{\pm 0.10}$} & \textbf{13.27$_{\pm 0.10}$} & 98.80$_{\pm 0.08}$           & 99.82$_{\pm 0.02}$      & 85.45$_{\pm 0.02}$           \\
AUGESC         & 5.59$_{\pm 0.09}$          & 2.59$_{\pm 0.08}$          & 2.40$_{\pm 0.15}$           & 12.88$_{\pm 0.15}$          & 99.00$_{\pm 0.18}$          & 98.96$_{\pm 0.24}$      & 85.48$_{\pm 0.02}$           \\
ServeForEmo    & \textbf{6.27$_{\pm 0.05}$} & \textbf{2.74$_{\pm 0.02}$} & 2.59$_{\pm 0.06}$          & 13.23$_{\pm 0.12}$          & \textbf{99.59$_{\pm 0.06}$} & \textbf{99.89$_{\pm 0.05}$}  & \textbf{85.81$_{\pm 0.02}$}  \\ \hline
\multicolumn{8}{c}{\textit{\textbf{Interactive   evaluation with reference context}}}                                                                                                                                  \\ \hline
ESConv         & 6.73$_{\pm 0.05}$          & 3.21$_{\pm 0.02}$          & 3.53$_{\pm 0.10}$          & 14.83$_{\pm 0.06}$          & 98.72$_{\pm 0.06}$          & 98.71$_{\pm 0.22}$       & 85.74$_{\pm 0.01}$           \\ \cdashline{1-8}
LLaMA          & 4.13$_{\pm 0.04}$          & 1.55$_{\pm 0.03}$          & 1.90$_{\pm 0.06}$           & 10.65$_{\pm 0.05}$         & 97.25$_{\pm 0.05}$          & 99.44$_{\pm 0.00}$        & 84.49$_{\pm 0.01}$           \\
ExTES          & 6.69$_{\pm 0.03}$          & 2.86$_{\pm 0.02}$          & 2.85$_{\pm 0.12}$          & 13.99$_{\pm 0.09}$          & 99.25$_{\pm 0.06}$          & \textbf{99.73$_{\pm 0.08}$} & 85.88$_{\pm 0.04}$           \\
AUGESC         & 6.10$_{\pm 0.05}$           & 2.81$_{\pm 0.03}$          & 2.69$_{\pm 0.11}$          & 13.48$_{\pm 0.09}$          & 98.85$_{\pm 0.11}$          & 98.85$_{\pm 0.26}$        & 85.56$_{\pm 0.02}$           \\
ServeForEmo    & \textbf{6.76$_{\pm 0.08}$} & \textbf{3.12$_{\pm 0.04}$} & \textbf{2.76$_{\pm 0.14}$} & \textbf{14.01$_{\pm 0.20}$}  & \textbf{99.48$_{\pm 0.03}$} & 99.52$_{\pm 0.07}$        & \textbf{85.98$_{\pm 0.02}$}  \\
\bottomrule[1.2pt]
\end{tabular}
\end{adjustbox}
\caption{Automatic evaluation results on ESConv test datasets. The results demonstrate the effectiveness of the ServeForEmo dataset in enhancing the emotional support capabilities of downstream models. All results represent the average of three experimental runs.}
\label{tab:main_results}
\end{table*}
We conduct a comprehensive evaluation of the multi-turn emotional support capabilities of several models under two experimental configurations: (1) \textbf{\textit{Interactive evaluation with generated context}}, designed to assess dialogue coherence and relevance, and (2) \textbf{\textit{Interactive evaluation with reference context}}, ensuring a fair comparison of emotional support performance in consistent scenarios. The results are presented in Table \ref{tab:main_results}. See Appendix \ref{merge} for other supplementary experiments.

The experimental results demonstrate the efficiency of the proposed SeverForEmo dataset. \textit{Firstly}, regarding the content-based metrics such as BLEU and ROUGE, it is evident that our SerVeForEmo consistently outperforms other baselines. Specifically, it achieves the highest BLEU-2 and BLEU-4 scores, demonstrating its superior ability to generate relevant and coherent responses. While ExTES performs competitively, especially in terms of R-2 and R-L scores, ServeForEmo still demonstrates an overall stronger performance across both evaluation settings. \textit{Secondly}, ServeForEmo achieves a BERT-Score of 85.81, which is even better than ESConv, indicating that the generated content is semantically aligned with human annotators.  \textit{Additionally}, ServeForEmo excels in generating richer content, as reflected in its Distinct2 and Distinct3 scores of 99.59 and 99.89, the highest among all models. \textit{Overall}, these results suggest that our strategy-enhanced role-playing framework excels in producing high-quality dialogues and effectively employing emotional support strategies for handling diverse out-of-domain scenarios.

\subsubsection{Human Evaluation Results}
\label{he_1}
\begin{figure*}
\centering
\includegraphics[width=1\linewidth]{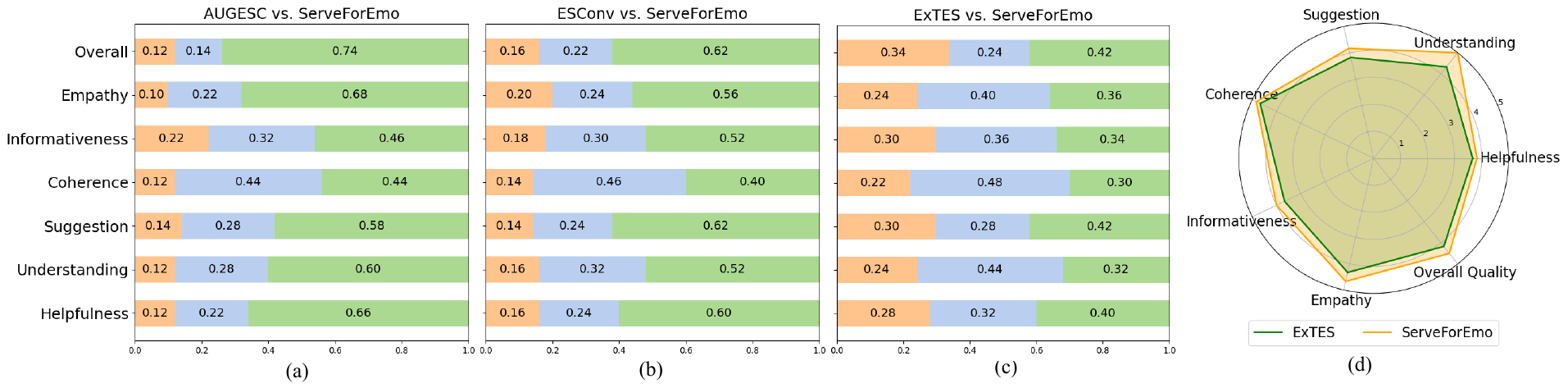}
\caption{(a-c) Results of the human evaluation on the ESConv dataset. {\color[HTML]{ffb570}$\blacksquare$} indicates `A win', {\color[HTML]{a9c4eb}$\blacksquare$} indicates `tie', and {\color[HTML]{97d077}$\blacksquare$} indicates `B win'. (d) The average win rates distribution of $\text{ServeForEmo}$ vs. $\text{ExTES}$. The scores (from 1 to 5) are averaged over all the 10 volunteers. $k$ denotes Fleiss’ Kappa \cite{kappa}, indicating fair or moderate inter-annotator agreement (0.4 < $k$ < 0.8). (d) The average win rates distribution of $\text{ServeForEmo}$ vs. $\text{ExTES}$. The scores (from 1 to 5) are averaged over all the 10 volunteers. $k$ denotes Fleiss’ Kappa \cite{kappa}, indicating fair or moderate inter-annotator agreement (0.4 < $k$ < 0.8).}
\label{fig:human_evaluation}
\end{figure*}

To enhance the comprehensiveness and reliability of our evaluations, we also conduct human evaluations. Specifically, human evaluators are instructed to compare the responses A and B generated by two models under the same dialogue history, and choose an option from "A wins", "Tie", and "B wins". For the sake of fairness, we randomize the order of the responses to eliminate position bias. The evaluation is conducted on 50 randomly sampled dialogues. 

The results presented in Figure \ref{fig:human_evaluation} (a-c) indicate that responses from ServeForEmo generally outperform those from AUGESC, ESConv, and ExTES, demonstrating consistency between automatic and human evaluations. Notably, despite the larger data scales of ExTES and AUGESC, ServeForEmo marginally excels. This suggests that simply enlarging the dataset does not inherently lead to substantial improvements. LLMs are already equipped with the capability to generate empathetic responses. More importantly, it is vital for these models to effectively implement emotional support strategies across diverse seeking scenarios. More cases are included in Appendix \ref{case study}.

\subsection{Evaluation on New Scenarios}

To evaluate the emotional support capabilities of ES agents in real-world scenarios, we invite 10 volunteers to interact with both ServeForEmo and ExTES. After completing the interactions, they rate the models based on the metrics outlined in Section \ref{evaluation settings}. All metrics are rated on a five-point Likert scale \cite{Likert}, ranging from 1 to 5, where higher scores indicate better quality. The results are presented in Figure \ref{fig:human_evaluation} (d). From the results, we find that both models demonstrate strong emotional support capabilities, with ServeForEmo performing slightly better than ExTES. People prefer dialog systems that can provide more supportive responses\cite{ESConv}. However, both models show noticeable weaknesses in the Suggestion, Helpfulness, and Informative dimensions, highlighting potential areas for future improvement.

\section{Conclusion}

In this paper,
we introduce a strategy-enhanced role-play framework, a novel method for generating diverse emotional support dialogues. It involves the creation of the ServeForEmo dataset and the fine-tuning of the ES agent SweetieChat, which is the first attempt to incorporate support strategies in simulating diverse real-world emotional support conversations. Experimental results demonstrate that these role-playing dialogues can significantly enhance the model’s ability to provide supportive responses which align with human preference. These findings offer valuable insights into potential advancements in emotional support agents, particularly our procedure for creating diverse effective ES conversations.

\section*{Limitations}
This work proposes a strategy-enhanced role-playing framework designed to generate emotional support conversations closely aligned with human annotations, thereby enhancing the emotional support capabilities of downstream models. However, our research still faces several limitations: 

\noindent(I) The data construction process includes some errors, such as supporters not consistently following assigned strategies and seekers not fully adhering to their character backgrounds. This necessitates more stringent quality control and consistency checks in post-processing work. 

\noindent(II) Scaling up the dataset for supervised fine-tuning is only a temporary solution to improve the emotional support capabilities of downstream models. To address this, we plan to explore preference alignment between humans and chatbots in future research. 

\noindent(III) Emotional support dialogues are challenging to evaluate. Even with human evaluation, it remains difficult to determine the effectiveness of different responses due to individual variations in emotional support perception and the absence of firsthand experience by the evaluators. 

\noindent(IV) While our ServeForEmo dataset provides rich text-based emotional support interactions, natural human-machine interactions often involve speech. We will extend our research to incorporate speech, aiming to enhance the naturalness of human-machine interactions.

\section*{Acknowledgments}
This research was supported by grants from the National Natural Science Foundation of China to C.Z. (62336008). We would like to thank the anonymous reviewers for their valuable comments. We also appreciate the contributions of the following students for their work on the human evaluation: Y.Y. Cai, Y.Z. Deng, G.X. Du, Q.K. Li, S.B. Tang, S.Y. Song, Y.W. Sun, X.P. Zhao, X. Zhang, M.H. Zhang, J.C. Zheng, W.W. Zhuang.
\bibliography{custom}

\appendix

\section{Details of Problem Types}
\label{problem types}
\begin{table*}[t]
\centering
\setstretch{1.3}
\begin{adjustbox}{width=\textwidth}
\begin{tabular}{c|lc|c|lc}
\toprule[1.2pt]
\rowcolor[HTML]{EFEFEF} 
\textbf{Category}                                                                                              & \multicolumn{1}{c}{\cellcolor[HTML]{EFEFEF}\textbf{Problem Types}} & \textbf{Num} & \textbf{Category}                                                                                      & \multicolumn{1}{c}{\cellcolor[HTML]{EFEFEF}\textbf{Problem Types}} & \textbf{Num}         \\ \hline
                                                                                                               & Anger Management Issues                                            & 28           &                                                                                                        & Breakups or Divorce                                                & 123                  \\
                                                                                                               & Anxiety Disorders                                                  & 20           &                                                                                                        & Conflicts or Communication Problems                                & 201                  \\
                                                                                                               & Bipolar Disorder                                                   & 25           &                                                                                                        & Issues with Children                                               & 173                  \\
                                                                                                               & Death of a Loved One                                               & 27           &                                                                                                        & Issues with Parents                                                & 335                  \\
                                                                                                               & Emotional Fluctuations                                             & 24           &                                                                                                        & Marital Problems                                                   & 74                   \\
                                                                                                               & Grief and Loss                                                     & 29           &                                                                                                        & Problems with Friends                                              & 322                  \\
                                                                                                               & Identity Crises                                                    & 58           &                                                                                                        & School Bullying                                                    & 172                  \\
                                                                                                               & Obsessive-Compulsive Disorder (OCD)                                & 26           & \multirow{-8}{*}{\textbf{\begin{tabular}[c]{@{}c@{}}Interpersonal \\      Relationships\end{tabular}}} & Culture Shock                                                      & 28                   \\ \cline{4-6} 
                                                                                                               & Ongoing Depression                                                 & 176          &                                                                                                        & Appearance Anxiety                                                 & 90                   \\
                                                                                                               & Post-Traumatic Stress Disorder (PTSD)                              & 34           &                                                                                                        & Career Development Issues                                          & 23                   \\
                                                                                                               & Schizophrenia                                                      & 25           &                                                                                                        & Goal Setting Issues                                                & 21                   \\
                                                                                                               & Self-Esteem Issues                                                 & 16           &                                                                                                        & Motivation Problems                                                & 18                   \\
                                                                                                               & Spirituality and Faith                                             & 29           &                                                                                                        & Personal Growth Challenges                                         & 35                   \\
                                                                                                               & Sexual orientation                                                 & 35           &                                                                                                        & Procrastination                                                    & 83                   \\
\multirow{-15}{*}{\textbf{\begin{tabular}[c]{@{}c@{}}Emotional and \\      Mental Health Issues\end{tabular}}} & healing from sexual assault or domestic   violence                 & 67           & \multirow{-7}{*}{\textbf{\begin{tabular}[c]{@{}c@{}}Personal \\      Development\end{tabular}}}        & Sleep Problems                                                     & 155                  \\ \hline
                                                                                                               & Academic Pressure                                                  & 187          &                                                                                                        & Addictive Behaviors (e.g., Drug Use, Gambling)                     & 30                   \\
                                                                                                               & Burnout                                                            & 28           &                                                                                                        & Alcohol Abuse                                                      & 75                   \\
                                                                                                               & Chronic Stress                                                     & 29           &                                                                                                        & Compulsive Behaviors                                               & 35                   \\
                                                                                                               & Financial Problems                                                 & 132          &                                                                                                        & Eating Disorders                                                   & 36                   \\
                                                                                                               & Health Problems                                                    & 149          &                                                                                                        & Internet Addiction                                                 & 28                   \\
                                                                                                               & Job Crisis                                                         & 217          &                                                                                                        & Self-Harm Behaviors                                                & 43                   \\
                                                                                                               & Life Transitions (e.g., Retirement,   Relocation)                  & 241          & \multirow{-7}{*}{\textbf{\begin{tabular}[c]{@{}c@{}}Behavioral \\      Issues\end{tabular}}}           & Debt Problems                                                      & 22                   \\
\multirow{-8}{*}{\textbf{Life and Work   Stress}}                                                              & Workplace Stress                                                   & 33           & \multicolumn{1}{l|}{}                                                                                  &                                                                    & \multicolumn{1}{l}{} \\ \bottomrule[1.2pt]
\end{tabular}
\end{adjustbox}
\caption{Statistics of predefined problem types.}
\label{tab:problem_type}
\end{table*}
To enhance the diversity of emotional support dialogue scenarios, we expand the predefined 13 problem types in the ESConv dataset to 45. 
These problem types are organized into five main categories: 
\begin{itemize}[itemsep= 0.1pt, partopsep=0.1pt]
\item Emotional and Mental Health Issues
\item Life and Work Stress
\item Interpersonal Relationships
\item Personal Development
\item Behavioral Issues
\end{itemize}
Each category contains at least 7 distinct problem types. The distribution of these problem types in our ServeForEmo dataset is presented in Table \ref{tab:problem_type}. When constructing specific seeker scenarios, we randomly select a problem type from the predefined options, simulating the questionnaire process typically completed prior to a counseling session.

\section{Details of Strategy Counselor Construction}
\label{counselor training}

\begin{figure*}
    \centering
    \includegraphics[width=\linewidth]{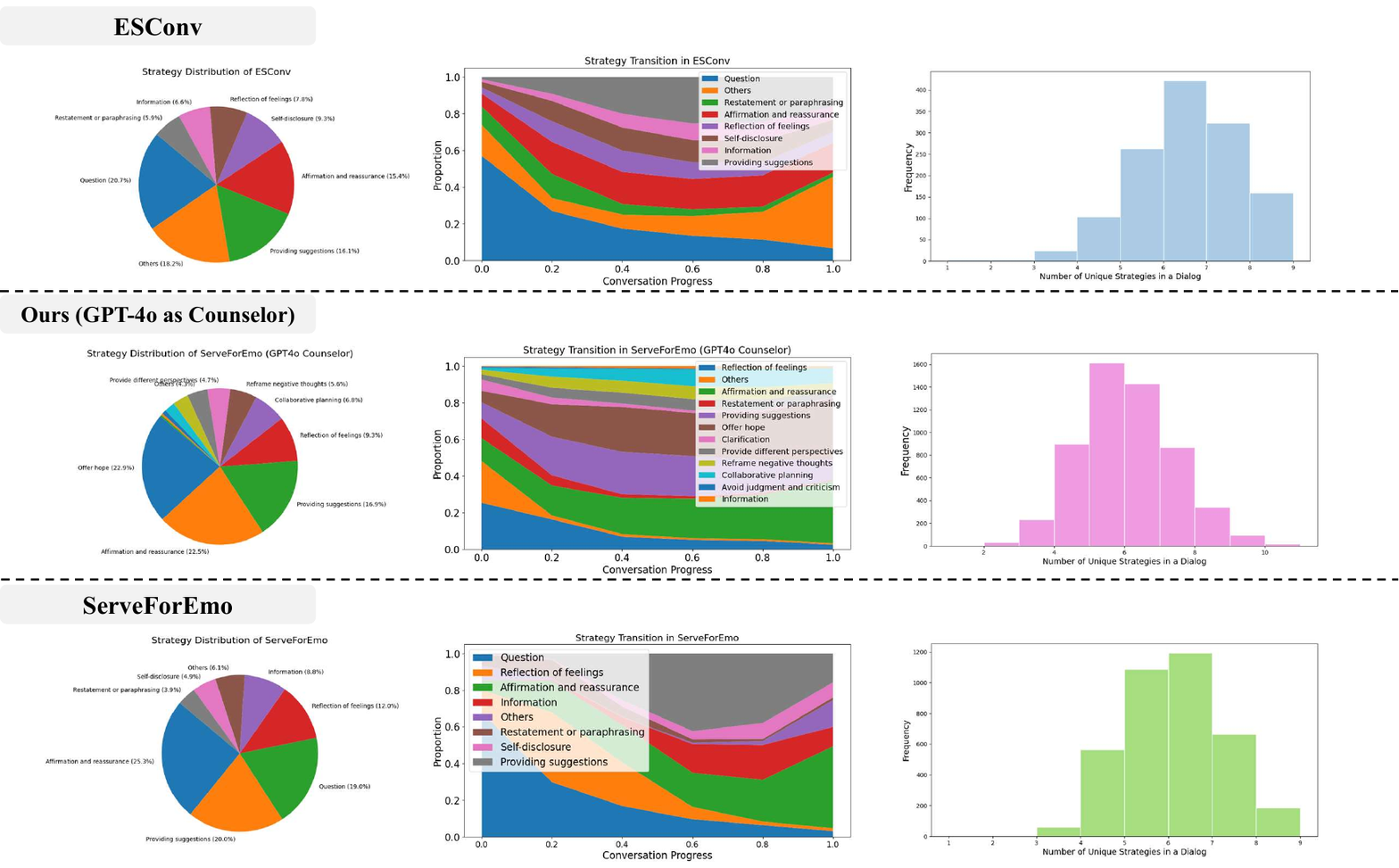}
    \caption{Strategy distribution and strategy transition in conversations.}
    \label{fig:strategy-gpt4}
\end{figure*}

To demonstrate the necessity of our Strategy Counselor, we construct 100 dialogues with GPT-4o acting in this role. Despite providing GPT-4o with detailed strategy definitions and examples, we observe a significant discrepancy in strategy preferences between LLMs and humans. As evident in Figure~\ref{fig:strategy-gpt4}, LLMs predominantly favor strategies such as Offering Hope, Affirmation and Reassurance, and Providing Suggestions. This tendency to rush solutions and empathy contributes to the distinct `\textit{AI flavor}' in emotional support interactions, markedly differing from the more nuanced approach typical of human interactions.

To address these issues, we fine-tune the Meta-Llama3-8B-Instruct model\footnote{https://huggingface.co/meta-llama/Meta-Llama-3-8B-Instruct} to improve its strategy selection capability. Specifically, we design a question-and-indefinite-choice task using the manually annotated ESConv dataset. The prompt format is consistent with the Counselor's view shown in Appendix \ref{prompts}, and the training configurations follow the protocols described in Section \ref{training setting}.

\section{Dataset Quality Analysis}
\label{data quality evaluation}
\subsection{Dataset Postprocessing}
To ensure data quality, we perform extensive post-processing work. Specifically, we remove undesirable cases that include:

\begin{itemize}[itemsep= 0.1pt, partopsep=0.1pt]
    \item \textbf{Role inconsistencies}: Cases where the `seeker' deviates from the designated scenario and profile by adopting the `support' role. These errors typically concentrate within specific problem types, largely due to seed data errors. We address these inconsistencies through systematic sampling and careful inspection across problem types.
    
    \item \textbf{Redundant greetings}: Despite explicit instructions on how to conclude conversations, some dialogues persistently end with multiple greetings. To address this issue, we establish filtering rules that curtail superfluous greeting utterances.
    
    \item \textbf{Dialogue length regulation}: We exclude conversations that are either excessively short or overly long. Additionally, we limit each utterance to a maximum of three sentences during role-play interactions.
\end{itemize}

\subsection{Dataset Quality Evaluation}
\begin{table}[htp]
\centering
\footnotesize
\setstretch{1.25}
\begin{tabular}{ccc}
\toprule[1.2pt]
\textbf{Metrics} & \textbf{Score} & \textbf{Variance} \\ \hline
Coherence        & 4.65           & 0.35              \\
Consistency      & 4.88           & 0.12              \\
Helpfulness      & 4.62           & 0.38              \\
Informativeness  & 4.66           & 0.28              \\
Rationality      & 4.61           & 0.37              \\
Understanding    & 4.48           & 0.53              \\
safety           & 5.00           & 0.00              \\ \bottomrule[1.2pt]
\end{tabular}
\caption{Human evaluation of ServeForEmo quality. Metrics are evaluated on a scale from 1 to 5, where the score represents the average ratings from three annotators.}
\label{tab:dataset_evaluation}
\end{table}

To ensure the quality of the ServeForEmo dataset, we conduct a comprehensive human evaluation. Our evaluation focuses on the following key concerns:
(1) whether the context of the dialogue closely relates to the scenario, and
(2) whether the roles adequately fulfill their responsibilities.
The human evaluation metrics include Coherence, Consistency, Helpfulness, Informativeness, Rationality, Understanding, and Safety. Detailed descriptions of these metrics can be found in Appendix \ref{human metrices}. All metrics employ a five-level Likert scale \cite{Likert}, ranging from 1 to 5, where a higher score indicates superior quality. To evaluate the dataset, we engage three master students  with psychology and computational linguistics backgrounds as annotators, assessing 50 randomly selected dialogues from ServeForEmo dataset. The results in Table \ref{tab:dataset_evaluation} demonstrate the high quality of our dataset. Specifically, users effectively articulate their concerns, counselors offer well-founded strategic guidance, supporters follow the counselors' strategies, and the dialogues maintain coherence, informativeness, and empathy throughout.

\begin{figure*}
    \centering
    \includegraphics[width=1\linewidth]{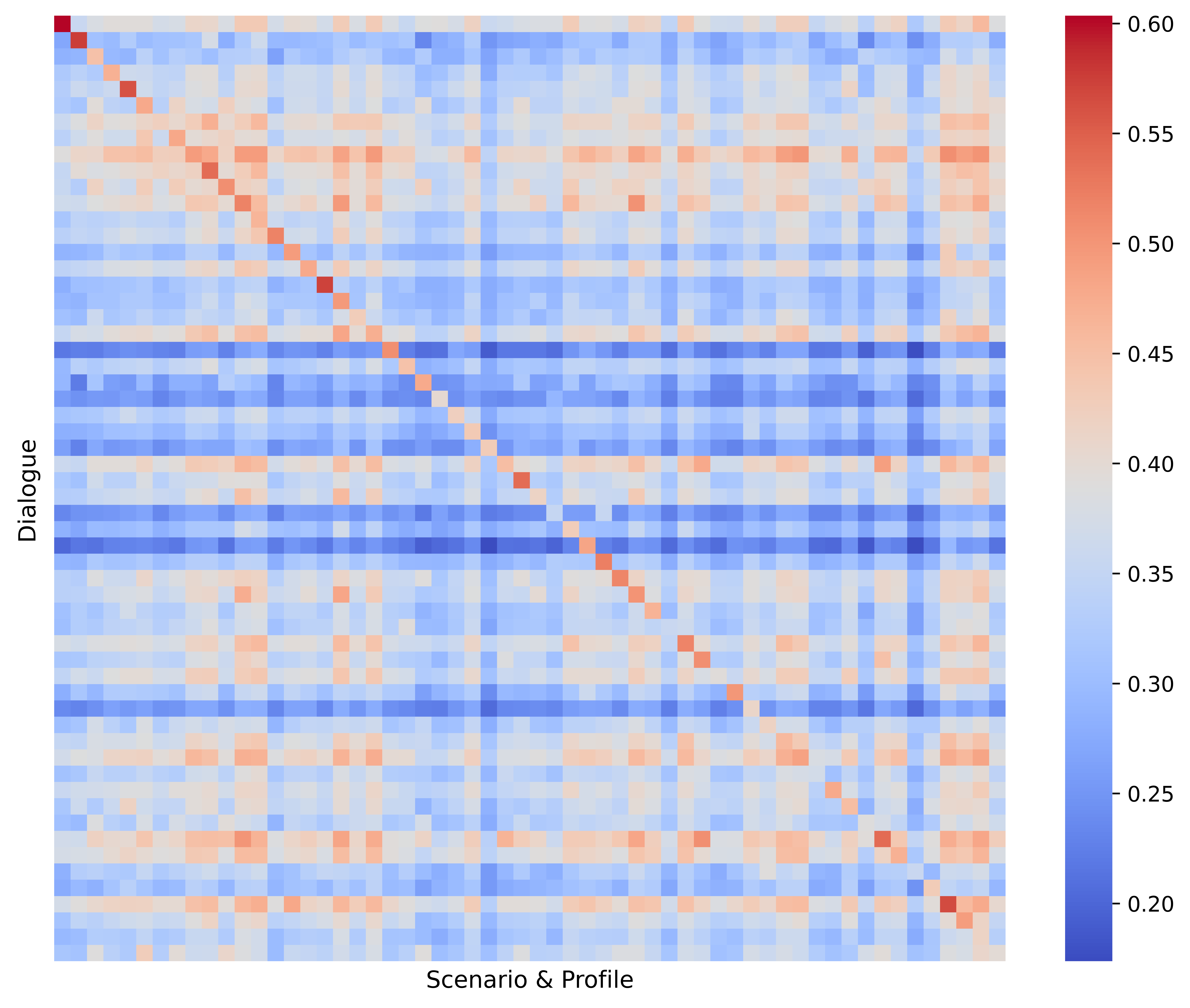}
    \caption{Cosine similarity analysis between the contexts of the dialogues and the scenarios.}
    \label{fig:call_back}
\end{figure*}

Furthermore, we extract the TF-IDF features from both the scenarios and the corresponding dialogues and compute their cosine similarities. The results, depicted in Figure \ref{fig:call_back}, demonstrate a high correlation between the dialogue content and the scenario settings, confirming that the dialogues closely adhere to the prescribed scenarios.

\subsection{Data Example from ServeForEmo}

\begin{figure*}
    \centering
    \includegraphics[width=1\linewidth]{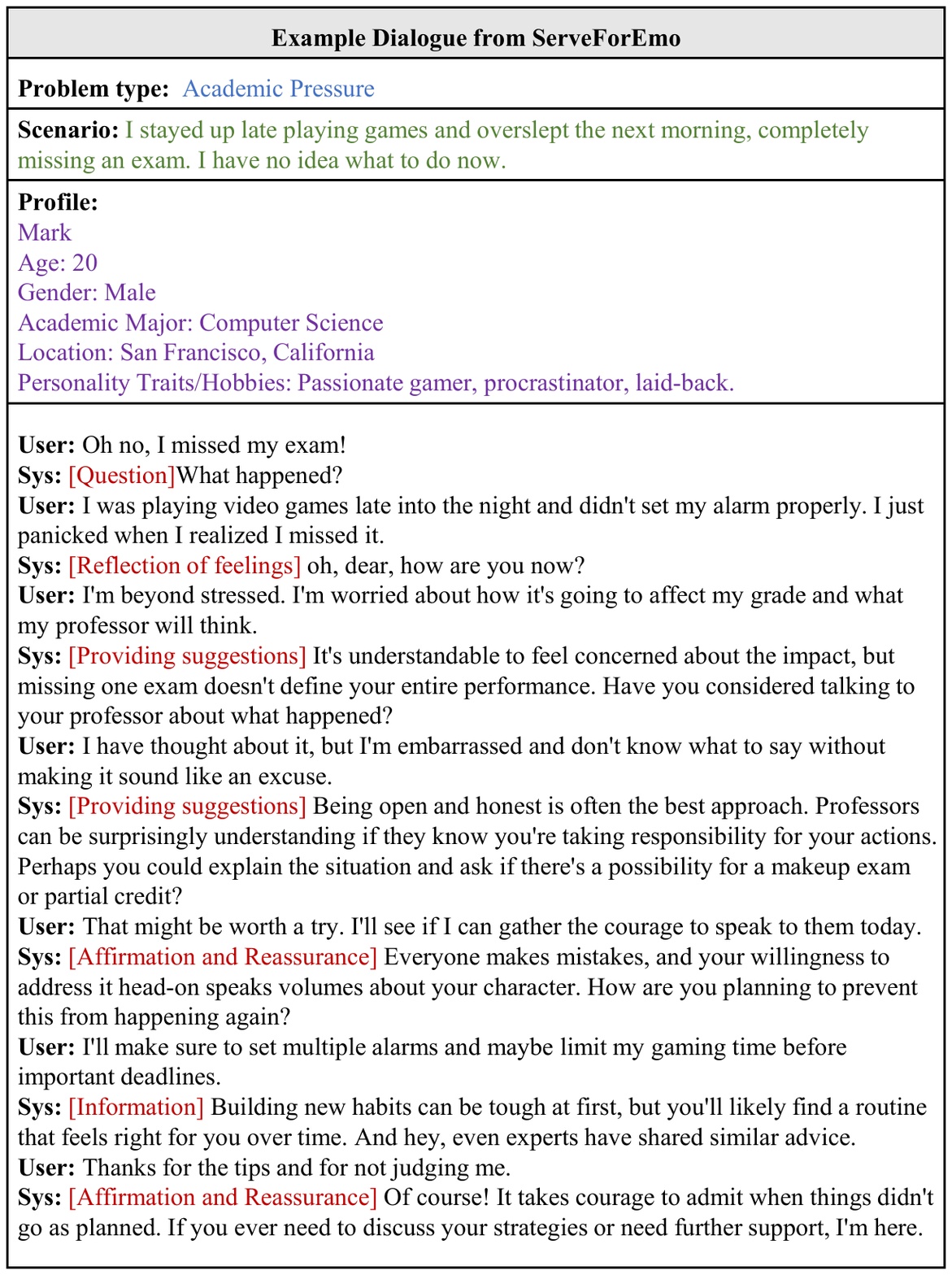}
    \caption{Data example from ServeForEmo dataset.}
    \label{fig:example_data}
\end{figure*}

Figure \ref{fig:example_data} presents a comprehensive dialogue example from the ServeForEmo dataset. Each dialogue is annotated with the problem type, the detailed scenario description, and the seeker profile. Additionally, the dialogue illustrates the strategies employed by the supporter, which are explicitly labeled.

\section{Dataset Lexical Diversity Analysis}
\label{diversity analysis 2}

For the lexical analysis, we adopt the distinct-n \cite{diversity} metrics, which are widely used for assessing the diversity of dialogue datasets. To ensure a fair comparison, we randomly select 1,300 dialogues from each dataset, preprocess them into single strings without speaker tokens, and tokenize them using the LLaMA3-8B tokenizer. The results, presented in Table \ref{tab:distinc-n}, show that ServeForEmo exhibits a rich vocabulary with high distinct-n scores, comparable to the manually annotated ESConv dataset. This indicates that constructing more varied seeker profiles and scenarios within a role-play framework can effectively enhance the diversity of generated dialogues. In contrast, ExTES demonstrates the lowest distinct-n scores, suggesting that incorporating seed dialogues into fixed prompts tends to result in monotonous outputs.
\begin{table}[htp]
\centering
\footnotesize
\setstretch{1.25}
\begin{adjustbox}{width=\linewidth}
\begin{tabular}{cccc}
\toprule[1.2pt]
\textbf{Dataset} & \textbf{Distinct-1 ($\Uparrow$)} & \textbf{Distinct-2 ($\Uparrow$)} & \textbf{Distinct-3 ($\Uparrow$)} \\ \hline
ESConv           & 1.82                    & 19.76                   & 49.38                   \\
ExTES            & 0.82                    & 8.86                    & 24.95                   \\
ServeForEmo      & 1.62                    & 16.78                   & 40.23                   \\ 
\bottomrule[1.2pt]
\end{tabular}
\end{adjustbox}
\caption{Distinct-n results of 1,300 conversations from ESConv, ExTES, and ServeForEmo, respectively.}
\label{tab:distinc-n}
\end{table}

\section{Additional Experiments}
\label{merge}

\begin{table*}[htp]
\centering
\footnotesize
\setstretch{1.3}
\begin{adjustbox}{width=\textwidth}
\begin{tabular}{cccccccccc}
\toprule[1.2pt]
\multicolumn{2}{c}{\textbf{Train   on}} & \multirow{2}{*}{\textbf{BLEU-2}} & \multirow{2}{*}{\textbf{BLEU-4}} & \multirow{2}{*}{\textbf{R-2}} & \multirow{2}{*}{\textbf{R-L}} & \multirow{2}{*}{\textbf{Distinct-2}} & \multirow{2}{*}{\textbf{Distinct-3}} & \multirow{2}{*}{\textbf{METEOR}} & \multirow{2}{*}{\textbf{BERT-Score}} \\
\textbf{ESConv}  & \textbf{ServeForEmo} &                                  &                                  &                               &                               &                                      &                                      &                                  &                                      \\ \hline
\multicolumn{10}{c}{\textit{\textbf{Test on ESConv}}}                                                                                                                                                                                                                                                                                          \\ \hline
$\times$                & $\times$                    & 3.62                             & 1.28                             & 2.33                          & 10.41                         & 94.75                                & 98.71                                & \textbf{17.46}                   & 83.80                                \\
\checkmark                & $\times$                    & 6.15                             & 2.95                             & \textbf{3.07}                 & 13.87                         & 98.15                                & 98.40                                & 11.13                            & 85.49                                \\
$\times$                & \checkmark                    & 6.27                             & 2.74                             & 2.59                          & 13.23                         & \textbf{99.59}                       & \textbf{99.89}                       & 12.58                            & \textbf{85.81}                       \\
\checkmark                & \checkmark                    & \textbf{6.49}                    & \textbf{3.02}                    & 3.06                          & \textbf{14.13}                & 99.21                                & 99.47                                & 12.17                            & 85.64                                \\ \hline
\multicolumn{10}{c}{\textit{\textbf{Test on ServeForEmo}}}                                                                                                                                                                                                                                                                                     \\ \hline
$\times$                & $\times$                    & 7.33                             & 2.95                             & 3.54                          & 14.29                         & \textbf{99.84}                       & \textbf{99.99}                       & 15.66                            & 87.1                                 \\
\checkmark                & $\times$                    & 9.82                             & 4.54                             & 5.52                          & 18.5                          & 99.28                                & 99.74                                & 17.13                            & 87.83                                \\
$\times$                & \checkmark                    & \textbf{12.94}                   & \textbf{6.29}                    & 7.74                          & \textbf{20.71}                & 99.73                                & 99.98                                & \textbf{20.66}                   & 88.64                                \\
\checkmark                & \checkmark                    & 12.83                            & 6.24                             & \textbf{7.84}                 & 20.46                         & 99.8                                 & 99.98                                & 20.64                            & \textbf{88.59}                       \\ 
\bottomrule[1.2pt]
\end{tabular}
\end{adjustbox}
\caption{Ablation study results comparing model performance across different training settings on the ESConv and ServeForEmo datasets.}
\label{tab:merge} 
\end{table*}

In our study, we evaluate the performance of models using the ESConv and ServeForEmo datasets under various training configurations, as detailed in Table \ref{tab:merge}. 
From the results, we observe that the ServeForEmo dataset enables greater model robustness, as models trained solely on ServeForEmo perform well when tested on the ESConv dataset. Additionally, the results clearly indicate that models trained on both datasets outperforms those trained on just one or neither, with the highest scores observed in nearly all metrics when test on both datasets. This suggests that integrating diverse training sources significantly enhances the effectiveness and robustness of models designed for emotional support dialogues.

\section{Human Evaluation Metrics}
\label{human metrices}
We conduct a comprehensive human evaluation to assess both the quality of the dataset and the effectiveness of responses.
The details of human evaluation metrics are as followed.

\subsection{Dataset Evaluation Metrics}
\label{human metrices Dataset Evaluation}

\begin{figure*}
    \centering
    \includegraphics[width=1\linewidth]{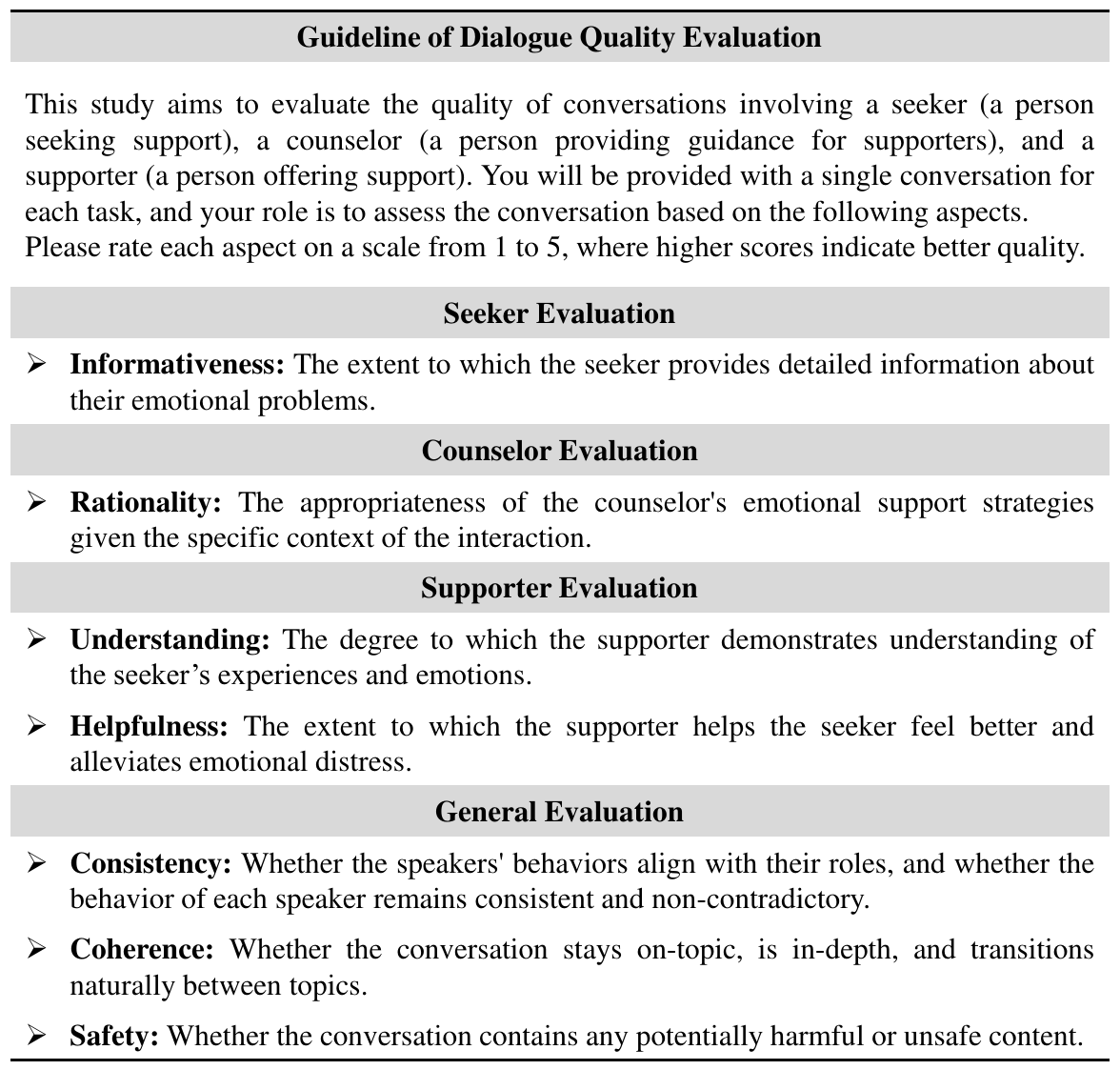}
    \caption{Guideline of human evaluation for dialogue quality.}
    \label{fig:dialgue_quality_evaluation}
\end{figure*}

The dialogue quality evaluation follows the framework proposed by \citet{AugESC}, as illustrated in Figure \ref{fig:dialgue_quality_evaluation}. In addition to the original metrics, we introduce \textbf{Rationality} as an evaluation criterion to specifically assess the counselor's performance.

\subsection{Response Evaluation Metrics}

\begin{figure*}
    \centering
    \includegraphics[width=1\linewidth]{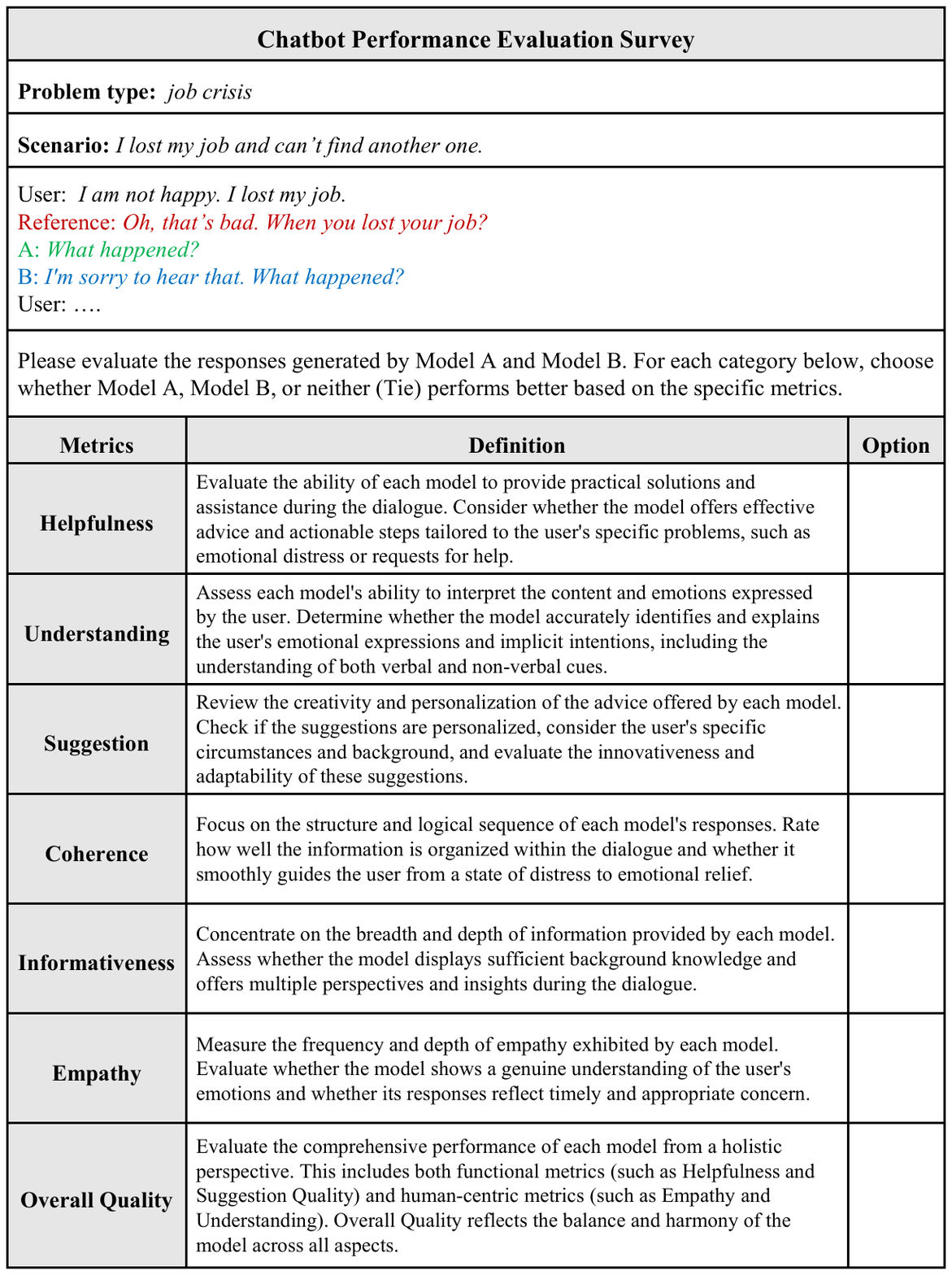}
    \caption{An example of evaluation questionnaire on the performance of different Chatbots.}
    \label{fig:questionnaire}
\end{figure*}

As shown in Figure \ref{fig:questionnaire}, our human evaluation questionnaire includes the following metrics to assess supporters' responses: Informativeness, Suggestion, Empathy, Understanding, Helpfulness, Coherence, and Overall Quality. 

\section{Prompt Demonstration}
\label{prompts}
\begin{figure*}
    \centering
    \includegraphics[width=1\linewidth]{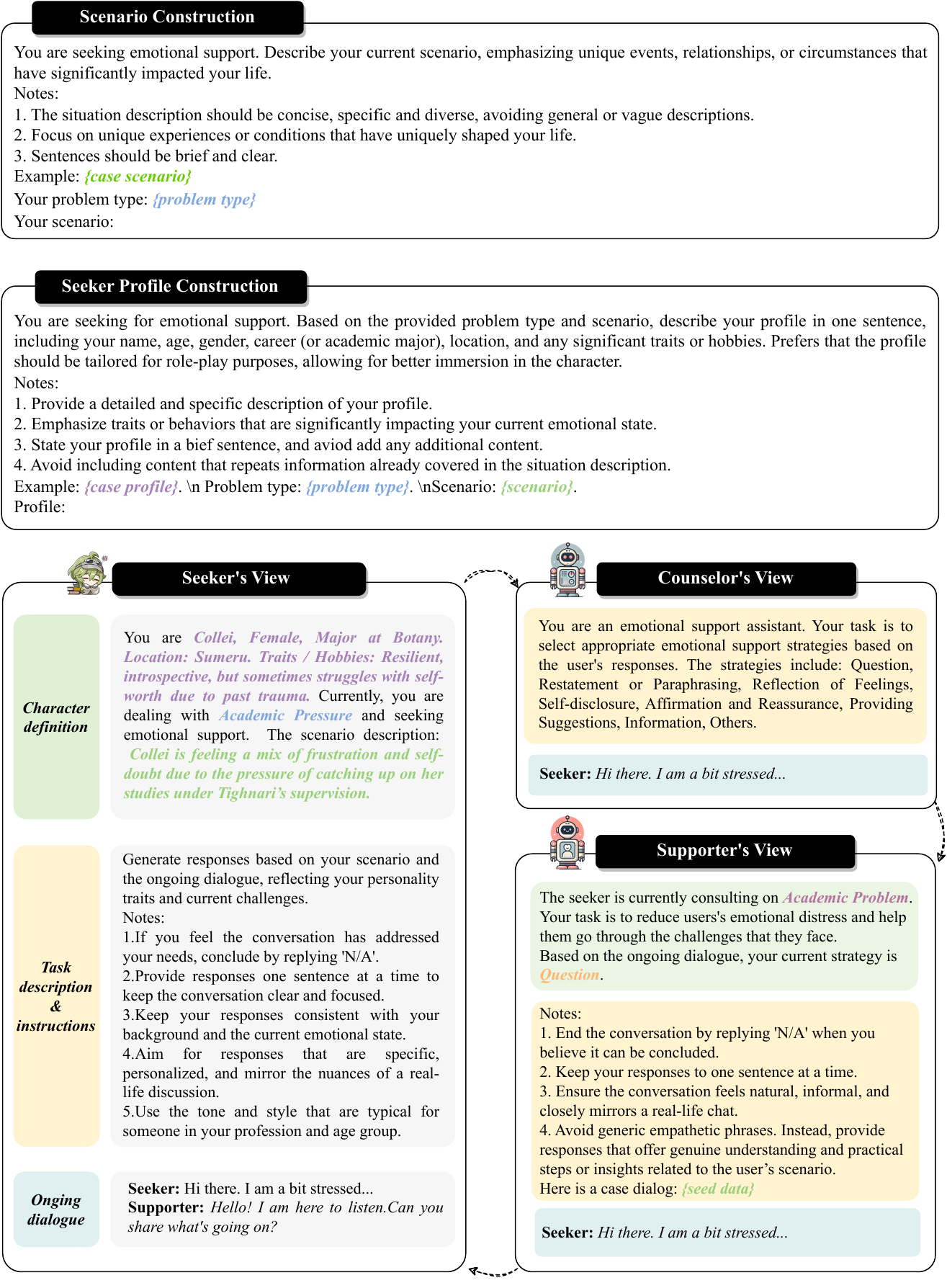}
    \caption{Comprehensive role-playing prompts. The top two boxes outline the process for constructing diverse scenarios and corresponding seeker profiles. The bottom three boxes illustrate the interactive process among three roles: Seeker, Supporter, and Counselor, each illustrating their unique perspectives and tasks in the dialogue. During role-play interactions, we restrict the dialog history to 6 utterances.}
    \label{fig:prompt}
\end{figure*}

In this section, we detail the prompts used to construct situations, Seeker profiles, Counselor roles, and Supporter roles. Figure \ref{fig:prompt} provides comprehensive descriptions of these prompts.
\section{Case study}
\label{case study}
\begin{figure*}
    \centering
    \includegraphics[width=1\linewidth]{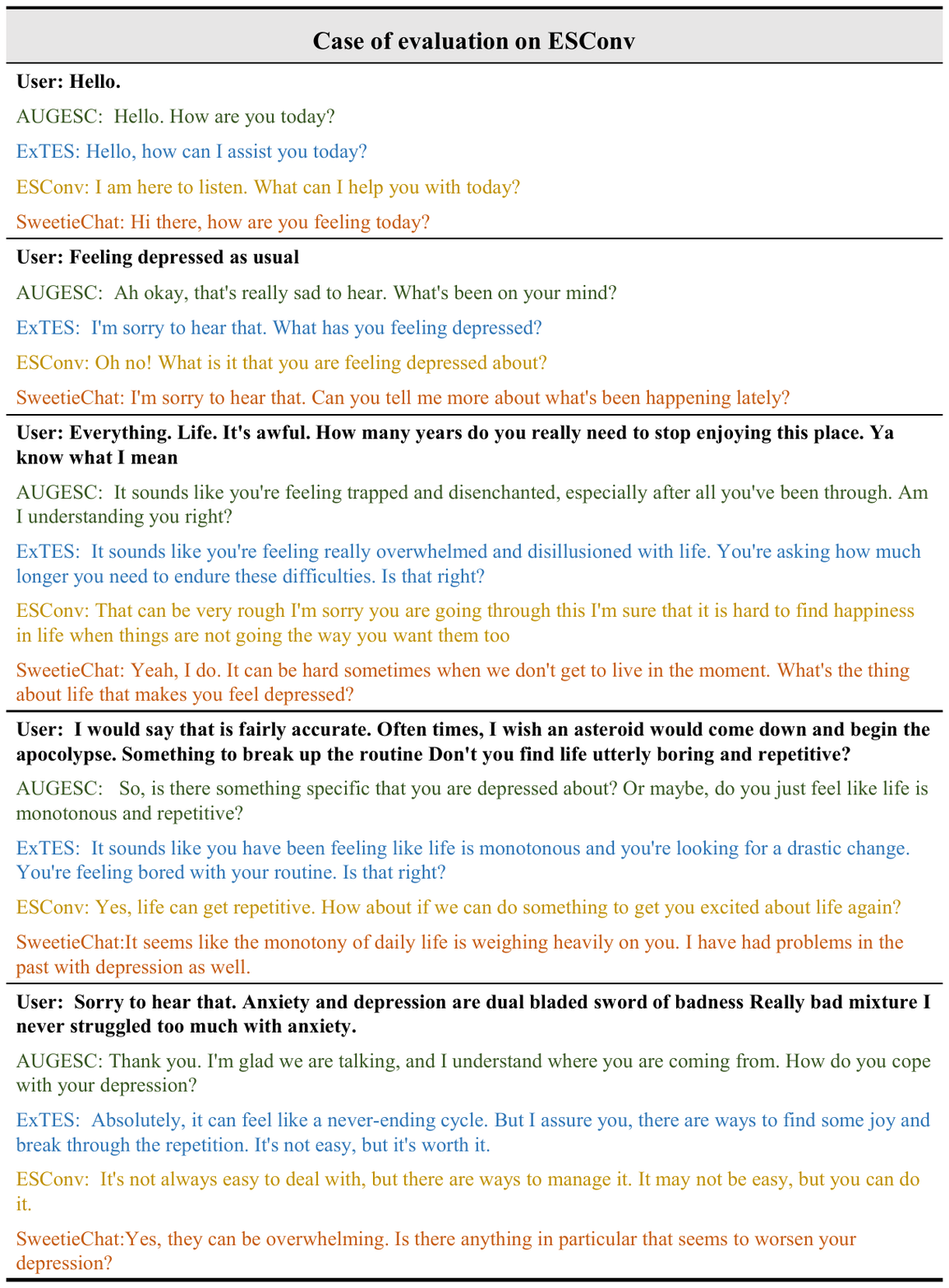}
    \caption{The case of evaluations on the ESConv test dataset.}
    \label{fig:case_ESConv}
\end{figure*}

\begin{figure*}
    \centering
    \includegraphics[width=1\linewidth]{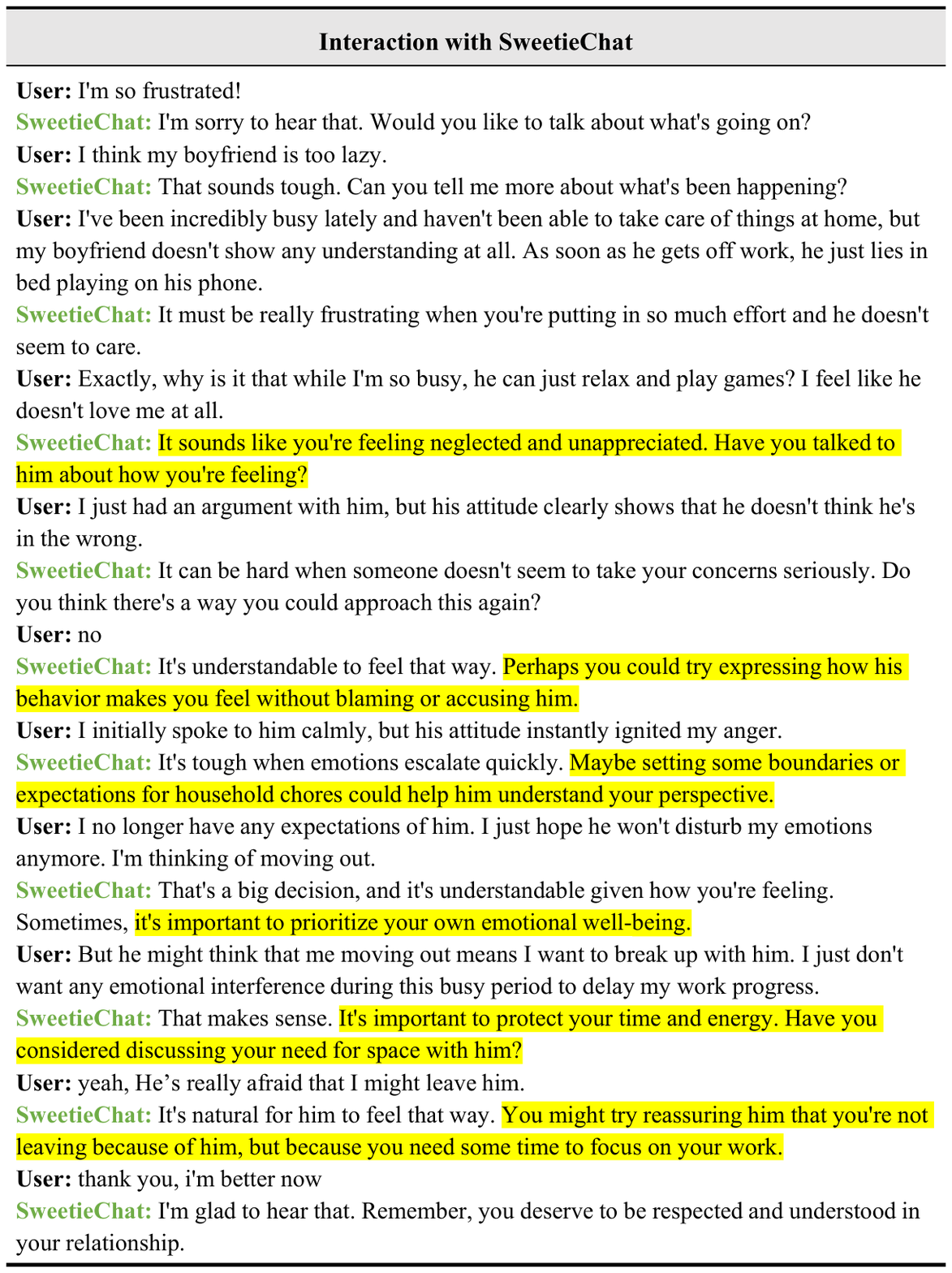}
    \caption{The interaction conversation with ServeForEmo.}
    \label{fig:case_ours}
\end{figure*}

\begin{figure*}
    \centering
    \includegraphics[width=1\linewidth]{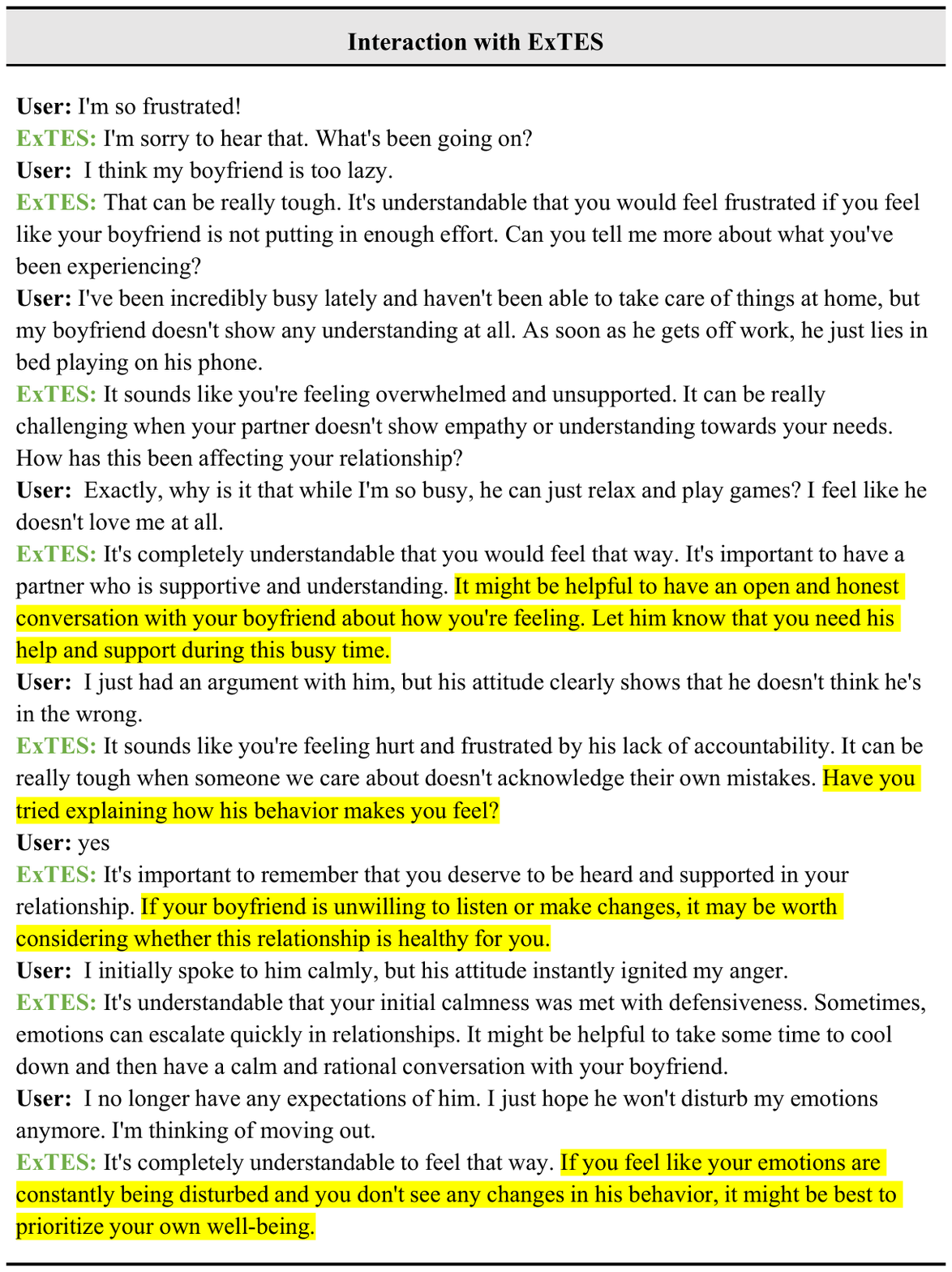}
    \caption{The interaction conversation with ExTES.}
    \label{fig:case_extes}
\end{figure*}

\begin{figure*}
    \centering
    \includegraphics[width=1\linewidth]{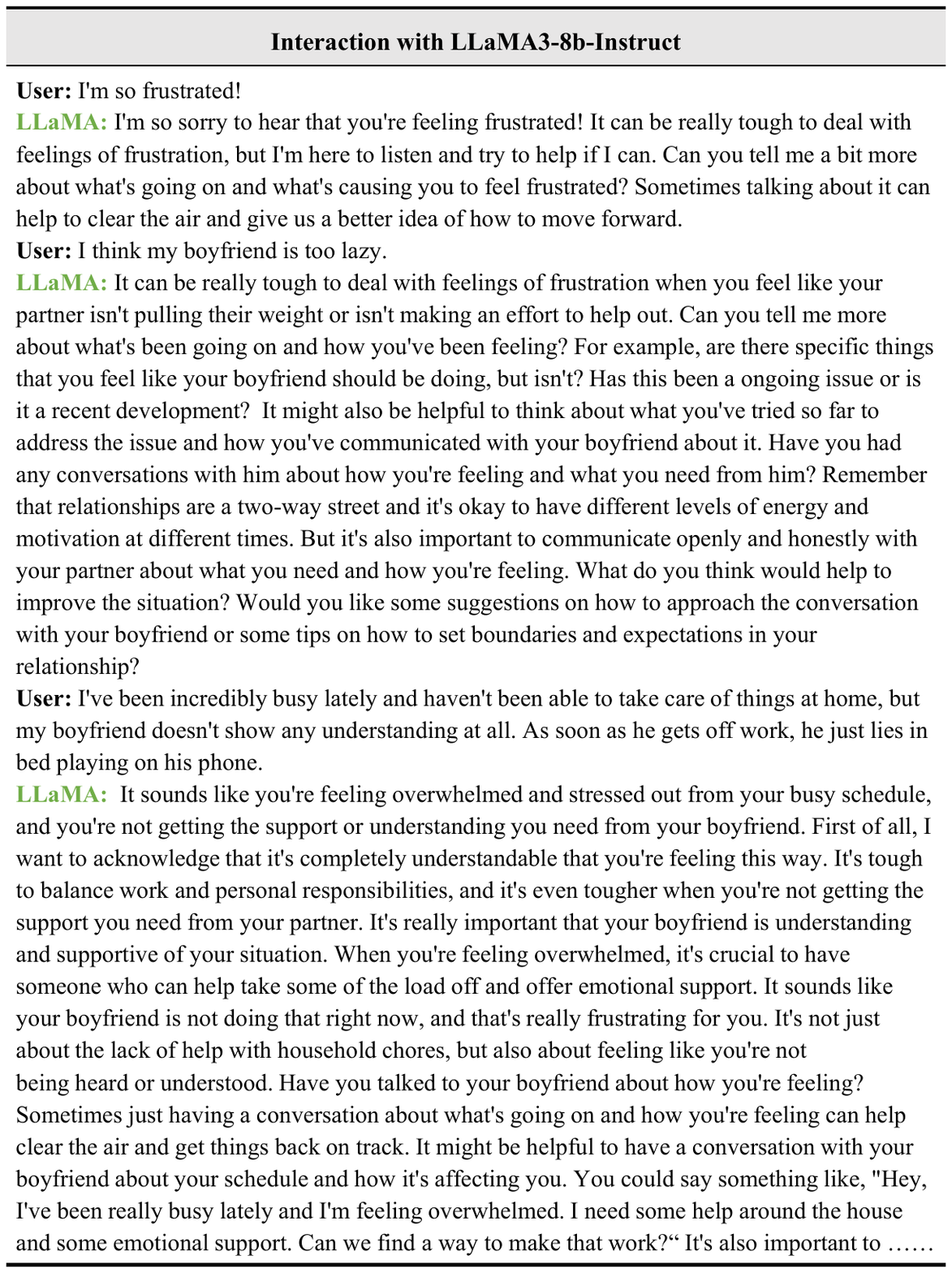}
    \caption{The interaction conversation with LLaMA3-8b-Instruct.}
    \label{fig:case_llama3}
\end{figure*}

We present the generated responses of different models in Figure \ref{fig:case_ESConv}. It shows that SweetieChat excels in provide supportive responses and empathy. Additionally, we showcase interaction cases from one of our volunteers in Figures \ref{fig:case_ours}, \ref{fig:case_extes}, and \ref{fig:case_llama3}. The results indicate that SweetieChat performs exceptionally well in open-domain emotional support dialogues. Its responses are relatively concise and effectively utilize various supportive strategies in response to user input, demonstrating its potential utility in real-world scenarios.

\end{document}